\documentclass[twoside,10pt]{article}
\usepackage{iclr2015,times} 
\iclrfinalcopy 
\usepackage{bm,mathtools,amsfonts}
\usepackage{enumerate}
\usepackage{graphicx}
\usepackage{caption}
\usepackage{subcaption}

\newcommand{\eqdef}{\overset{def}{=}}
\newcommand\x{{\mathbf x}}
\newcommand\btheta{{\bm \theta}}
\newcommand\y{{\mathbf y}}
\renewcommand\r{{\mathbf r}}
\newcommand\A{{\mathbf A}}
\newcommand\W{{\mathbf W}}
\renewcommand\H{{\mathbf H}}
\newcommand\M{{\mathbf M}}
\newcommand\J{{\mathbf J}}
\renewcommand\L{{\mathbf L}}
\newcommand\R{{\mathbf R}}
\newcommand\K{{\mathbf K}}
\newcommand\X{{\mathbf X}}
\renewcommand\O{{\mathbf O}}
\renewcommand\S{{\mathbf S}}
\newcommand\T{{\mathbf T}}
\newcommand\E{{\mathbf E}}
\newcommand\Y{{\mathbf Y}}
\newcommand\Z{{\mathbf Z}}
\newcommand\U{{\mathbf U}}
\newcommand\B{{\mathbf B}}
\newcommand\C{{\mathbf C}}
\newcommand\D{{\mathbf D}}
\newcommand\F{{\mathbf F}}
\newcommand\G{{\mathbf G}}
\newcommand\g{{\mathbf g}}
\newcommand\Q{{\mathbf Q}}
\newcommand\q{{\mathbf q}}
\newcommand\I{{\mathbf I}}
\def\Re{{\mathbb{R}}}
\def\tr{{\operatorname{tr}\,}}
\newcommand\diag{{\operatorname{diag}\,}}

\title{Parallel training of DNNs with Natural Gradient and Parameter Averaging}

\author{ Daniel Povey, Xiaohui Zhang \& Sanjeev Khudanpur \\ 
Center for Language and Speech Processing \& Human Language Technology Center of Excellence,\\
The Johns Hopkins University, Baltimore, MD 21218, USA \\
\texttt{\{dpovey@gmail.com\}, \{xiaohui,khudanpur@jhu.edu\}} \\
}

\begin{document}
\maketitle
\begin{abstract}
We describe the neural-network training framework used in the Kaldi speech
recognition toolkit, which is geared towards training DNNs with large amounts
of training data using multiple GPU-equipped or multi-core machines.  In order
to be as hardware-agnostic as possible, we needed a way to use multiple
machines without generating excessive network traffic.  Our method is to
average the neural network parameters periodically (typically every minute or
two), and redistribute the averaged parameters to the machines for further
training.  Each machine sees different data.  By itself, this method does not
work very well.  However, we have another method, an approximate and efficient
implementation of Natural Gradient for Stochastic Gradient Descent (NG-SGD),
which seems to allow our periodic-averaging method to work well, as well as
substantially improving the convergence of SGD on a single machine.
\end{abstract}

\section{Introduction}

Parallel training of neural networks generally makes use of some combination of
model parallelism and data parallelism~\citep{downpour}, and the normal approach
to data parallelism involves communication of model parameters for each
minibatch.  Here we describe our neural-net training framework which uses a
different version of data parallelism: we have multiple SGD processes on
separate machines, and only infrequently (every minute or so) average the model
parameters and redistribute them to the individual machines.  This is very
effective for us for large-scale training of systems for speech recognition-- but
in our case it only works well when combined with an efficient implementation
of natural gradient stochastic gradient descent (NG-SGD) that we have
developed.  We don't attempt in this paper to develop a framework that explains
why parameter averaging should work well despite non-convexity of DNNs, or why
NG-SGD is so helpful.  The point of this paper is to describe our methods and
to establish that empirically they work well.  The significance of this work is
that we show that it is possible to get a linear speedup when increasing the
number of GPUs, without requiring frequent data transfer (however, this only
holds up to about 4 or 8 GPUs).



In Section~\ref{sec:problem} we describe our problem setting, which
is Deep Neural Networks (DNNs) applied to speech recognition-- although our ideas
are more general than this.  In Section~\ref{sec:averaging} we introduce the
parallel training method.  In Section~\ref{sec:natural:gradient} we describe
the general ideas behind our natural gradient method, although most of the
technical details have been relegated to
appendices. 
In this paper we don't give any proofs, but we
do discuss in Section~\ref{sec:prove} what we think we can and can't
be proven about our methods.  Section~\ref{sec:experiments} has experiments
on the convergence of SGD with and without natural
gradient and parallelism.  We conclude in Section~\ref{sec:conc}.

There are two versions of our NG-SGD method: a ``simple'' version and an
``online'' one.  Technical details for these are in
Appendices~\ref{app:simple} and~\ref{app:online} respectively.
Appendix~\ref{appendix:other_aspects} has background information on our
DNN implementation.

\section{Problem setting}
\label{sec:problem}

 When training DNNs for speech recognition, the immediate problem 
 is that of classifying vectors $\x \in \Re^D$ as corresponding to discrete labels $y \in {\mathcal Y}$.
 The dimension $D$ is typically several hundred, with $\x$ being
 derived from short-time spectral properties of the acoustic signal; 
 and ${\mathcal Y}$ corresponds to clustered HMM states of
 context-dependent phones, $|{\mathcal Y}| \simeq 5000$ being typical.
 Each $(\x, y)$ pair corresponds to a single {\em frame} of speech data; frames
 are typically extracted at the rate of 100 per second, with a duration of 25 milliseconds,
 and $\x$ contains spectral information from several adjacent frames 
 spliced together~\citep{seide2011conversational}.
 We are ultimately not just interested in the top $y$ on each frame, but in the
 log-probabilities $\log p(y | \x)$ for all $y$, since we will use them
 as costs in a Viterbi search for a path corresponding to the most likely word sequence.
 The objective function for training is the sum, over all
 frames of training data, of the log-probability of $y$ given $\x$:
 $\sum_i \log p(y_i | \x_i)$.  Since we are maximizing this, our use of the
 term SGD is of course a slight misnomer; it is gradient {\em ascent}.
 The supervision labels $y$ are derived from
 a Viterbi alignment of a Hidden Markov Model (HMM) derived from the reference word 
 sequence of each training utterance.

\section{SGD with parameter averaging}
\label{sec:averaging}

 \subsection{Parameter-averaging overview}
 \label{sec:overview}

 The parameter-averaging aspect of our training is quite simple.
 We have $N$ machines (e.g. $N = 4$) each doing SGD
 separately with different randomized subsets of the training data, and
 we allow their parameters to gradually diverge.  After each machine has
 processed a fixed number of samples $K$ (typically $K = 400\,000$), we average
 the parameters across all the jobs and re-distribute the result to the
 machines.  (In practice we do this by spawning new jobs in GridEngine or in
 whatever job management system we are using).  This is repeated until we have
 processed all the data for a specified number of epochs, e.g. 10.

 We define an {\em outer iteration} of training as the time it takes for each
 job to process $K$ training examples.  The number of outer iterations
 per epoch depends on $K$ and the quantity of training data.

 We find it useful to define the effective learning rate of the parallel SGD
 procedure as the learning rate $\eta_t$ being used by the individual jobs,
 divided by the number of jobs $N$.  As we increase the number of jobs $N$, in
 order to get a linear speed up we need to increase the learning rate
 proportional to $N$ so that the effective learning rate stays the same.  The
 concept is that when we do the parameter averaging, the parameter update from any
 individual SGD job gets diluted $N$-fold.

 The reason why we set this up as parameter-averaging instead of summing
 the parameter changes from the respective jobs, is out of concern for stability.  Imagine there
 is some direction in parameter space where the Hessian is large enough (and our
 learning rate large enough) that stochastic gradient descent nearly reaches
 equilibrium after processing $K$ samples.  If there are, say, 4 jobs, then
 after processing $K$ examples and summing the parameter changes, the parameters
 end up not close to the equilibrium value but off in the opposite direction and
 3 times farther away than at the start.  It is clear that this would lead to
 divergence.

\subsection{Other aspects of our SGD implementation}

 At this point we provide some more details of other relevant
 features of our SGD implementation, namely the learning rate schedule and
 the way we enforce a maximum parameter change to prevent divergence.

 There are some other, less directly relevant issues which we have relegated to
 Appendix~\ref{appendix:other_aspects}: namely, CPU versus GPU-based SGD~(\ref{appendix:cpu:gpu});
 data randomization issues~(\ref{appendix:data:organize}); generalized model
 averaging~(\ref{sec:model_averaging}); mixture components
 a.k.a. sub-classes~(\ref{sec:mixture_components}); input data
 normalization~(\ref{sec:input_data_normalize}); parameter
 initialization~(\ref{sec:initialization}); sequence
 training~(\ref{sec:sequence_training}); and online decoding with
 i-vectors~(\ref{sec:online_decoding}).

 \subsubsection{Learning rate schedule}

 It was found in~\citep{senior2013empirical} that when training DNNs for speech
 recognition, an exponentially decreasing learning rate works well, and we
 independently found the same thing.  We generally use a learning rate that
 decreases by a factor of 10 during training, on an exponential schedule.
 Unless mentioned otherwise, for experiments reported here the learning rate
 starts at 0.01 and ends at 0.001.  We specify the number of epochs in
 advance; it is typically a number in the range 4 to 20 (if we have more
 data, we train for fewer epochs).

 Note that while in the experiments presented here we did not separately tune
 the learning rate schedule for SGD and NG-SGD (we just used values which had
 worked well in the past for NG-SGD), we have done extensive experiments in the
 past, on smaller setups, where the learning rate schedule was tuned
 independently; and in all circumstances we found NG-SGD to be helpful.  It was
 not feasible to repeat those experiments on a large-scale setup.

 \subsubsection{Maximum parameter change}
 \label{sec:max_param_change}

 A common pathology when doing SGD for deep learning is that during training,
 the parameters will suddenly start getting very large and the
 objective function will go to negative infinity.  This is known as parameter
 divergence.  The normal solution is to decrease the learning rate and 
 start the training again, but this is a very inconvenient.  To
 avoid this pathology, we modified the SGD procedure to enforce a maximum parameter change
 per minibatch.  This limit tends to be active only early in training, particularly for
 layers closer to the output.  We have provided further details on this
 in Appendix~\ref{appendix:weight_limit}.

 \section{Natural gradient for SGD}
 \label{sec:natural:gradient}

 In this section we describe our natural-gradient modification to SGD, in which
 we scale the gradients by a symmetric positive definite matrix that is an
 approximation to the inverse of the Fisher matrix.

 Technically speaking, Natural Gradient means taking a step along the gradient
 of a Riemannian parameter surface, which follows a curving path in conventional
 parameter space and which is extremely hard to compute.
 However, previous work~\citep{yang1998complexity,le2007topmoumoute} has used the term
 ``Natural Gradient'' to describe methods like ours which use an an approximated
 inverse-Fisher matrix as the learning rate matrix, so we follow their
 precedent in calling our method ``Natural Gradient''.  

 \subsection{We can replace the scalar learning rate in SGD with a matrix}
 
 In SGD, the learning rate is often assumed to be a scalar $\eta_t$, decreasing
 with time, and the update equation is something like
  $$  \btheta_{t{+}1} =   \btheta_t + \eta_t \g_t $$
 where $\g_t$ is the objective-function gradient sampled on time $t$ (e.g. computed from
 a training sample or a minibatch).  However, it is possible to replace this scalar with a symmetric
 positive definite matrix, and we can write instead:
\begin{equation}
\label{eqn:et}
    \btheta_{t{+}1} =   \btheta_t + \eta_t \E_t \g_t 
\end{equation}
 with $\E_t$ the matrix component of learning-rate; 
 it is more convenient for proofs to keep $\eta_t$ separate rather than absorbing it into
 $\E_t$.   It acceptable for $\E_t$ to be random: if we can bound the eigenvalues of $\E_t$
 above and below, by positive constants known in advance, and $\E_t$ and $\g_t$ are independently
 sampled given the parameter $\btheta$, then we can prove convergence under the same kinds of
 conditions as if we were using a scalar learning rate~\citep[Sec. 4.2.2]{bottou1998online}

 In general, the learning-rate matrix should not be a function of the data sample
 which we are currently processing, or it may prevent convergence to a local
 optimum.  As an example of this, a matrix that was systematically smaller for a
 particular type of training data would clearly bias the learning by downweighting
 that data.

 \subsection{The inverse Fisher matrix is a suitable learning-rate matrix}



 There are reasons from statistical learning theory, related
 to the Natural Gradient idea~\citep{Amari:1998:NGW:287476.287477}, why we may
 want to set the learning-rate matrix $\E_t$ to the inverse of the Fisher information matrix.
 See for example,~\citep{murata1999statistical} and~\citep{le2007topmoumoute}.
 The Fisher matrix is most directly defined for situations where we are 
 learning a distribution, as opposed to classification problems such as the current one.
 Suppose $x$, which may be discrete or continuous, is the variable whose distribution we are modeling,
 and $f(x; \btheta)$ is the probability or likelihood of $x$ given parameters $\btheta$,
 then the Fisher information matrix ${\cal I}(\btheta)$ is defined as the expected 
 outer product (second moment) of the derivative of the log-probability w.r.t. the parameters, i.e. of
 $$
  \frac{\partial}{\partial \btheta} \log f(x; \btheta) .
 $$
 This derivative is called the ``score'' in information theory.
 Part of the justification for this use of the Fisher matrix is that,
 under certain conditions, the Fisher matrix is identical to the Hessian;
 and it is obvious why the inverse Hessian would be a good gradient descent direction.
 These conditions are quite stringent, and include that the model is correct and $\btheta$ is
 at the value corresponding to the true data distribution; but even if these conditions do not
 apply, the Fisher information matrix is in some sense ``dimensionally'' the same as the Hessian-- that is,
 it transforms the same way under changes of parameterization-- so 
 its inverse may still be a good choice of learning-rate matrix.

 It is quite easy to generalize the notion of the Fisher matrix to a prediction task $p(y; x, \btheta)$.
 We write $p(y, x; \btheta) = q(x) p(y; x, \btheta)$ for a data distribution $q(x)$ that we assume to be 
 independently known (and not a function of $\btheta$).  
 It is not hard to see that the score equals just $\frac{\partial}{\partial \btheta} \log f(x; y, \btheta)$;
 since $q(x)$ does not depend on $\btheta$, there is no additional term involving $q(x)$.
 The expectation that we take when computing the Fisher matrix is taken over the joint distribution of $x$
 and $y$.  This argument also appears in~\citep[Section 3]{le2007topmoumoute}.

 Still more generally, we can compute a quantity that is analogous to Fisher matrix for
 any objective function, even one that does not represent a log-probability or
 log-likelihood; and we will still have a matrix that transforms in the same way as the
 Hessian under changes of variables - i.e. its inverse may still be a reasonable choice
 for a learning-rate matrix.

 \subsection{We need to approximate the Fisher matrix in practice}

 For large-scale problems, such as DNNs for speech recognition with millions of
 parameters, even one inversion of the Fisher matrix is impractical because it
 would take time $O(n^3)$ in the parameter dimension.  However, it may be
 practical to deal with factored forms of it.  There has been previous
 literature on this.  In~\citep{le2007topmoumoute}, the Fisher matrix was divided
 into diagonal blocks and each block was approximated by a low-rank matrix.
 The idea of diagonal blocks was also explored in~\citep{bastian2011simplified}, with
 one block per weight matrix; our approach uses the same idea.  In the
 unpublished manuscript~\citep{yang1997natural} (some of the material in which
 was later published in~\citep{yang1998complexity}), the authors attempted to
 show analytically that under certain quite strong assumptions, the Fisher
 matrix for a single-hidden-layer neural network has the form of a Kronecker
 product.  Although we are interested in
 more general networks than they considered, the Kronecker product does also
 appear in our factorization of the Fisher matrix.

 We should note that there are ways to use Natural Gradient without factorizing
 the Fisher information matrix, if one is willing to accept a significantly
 increased time per iteration.  See for
 example~\citep{DBLP:journals/corr/abs-1301-3584}, which uses a truncated Newton
 method to approximate multiplication by the inverse of the Fisher matrix.

 \subsection{Our factorization of the Fisher matrix}

 Our factored form of the Fisher matrix is as follows: given a
 neural network with $I$ weight matrices, we divide the Fisher 
 matrix into $I$ diagonal blocks, one for each weight matrix. 
 Consider the $i$'th diagonal block of the
 Fisher matrix, corresponding to the parameters of a weight matrix $\W_i$, and
 assume that there is no separate bias term (we can append a 1 to the inputs and include
 the bias term in the weight matrix).  The $i$'th block of the Fisher matrix is 
 a Kronecker product of two symmetric positive definite matrices: $\A_i$, whose dimension 
 is the output (row) dimension of $\W_i$, and $\B_i$, whose dimension is the input (column) dimension of 
 $\W_i$.  We further factorize the matrices $\A_i$ and $\B_i$ as
 a low-rank symmetric matrix plus a multiple of the identity matrix. 
 We write the approximated Fisher matrix in the form
\begin{equation}
 \F = \diag (\A_1 \otimes \B_1, \A_2 \otimes \B_2, \ldots, \A_I \otimes \B_I)
 \label{eqn:fisher:factorize}
\end{equation}
 where $\A_i$ and $\B_i$ are each factorized in the form $\lambda \I + \X \X^T$.
 The order in which $\A_i$ and $\B_i$ appear in the Kronecker product depends
 on the way in which we vectorize the weight matrices-- row-wise or column-wise.
 In practice we don't ever deal explicitly with these Kronecker products or
 vectorized weight matrices in the algorithm, so this choice doesn't matter.
 It is not hard to show that if the Fisher matrix can be factored this way,
 then its inverse can be factored the same way.

 \subsection{How we estimate the Fisher matrix}
\label{sec:a:b}

 We have two different methods for estimating the factorized Fisher matrix:
 \begin{itemize}
 \item  Simple method:
    We estimate the Fisher matrix from the {\em other} samples in the minibatch
    we are currently training on, holding out the current sample to avoid bias.  This can be done surprisingly 
    efficiently.  Details are in Appendix~\ref{app:simple}.
 \item Online method:
   We estimate the Fisher matrix from all previous minibatches,
   using a forgetting factor to downweight minibatches that are distant in
   time.  Details are in Appendix~\ref{app:online}.
 \end{itemize}
 We generally use the online method as it is significantly faster on GPUs and
 usually seems to lead to faster learning, probably due to the less noisy
 estimate of the Fisher matrix.  We describe the simple method because it is
 easier to understand and helps to motivate the online method.

 \subsection{Operation on vectors}

 Although we describe our Fisher matrix as as a Kronecker product, we do not
 have to explicitly construct this product in our code.

 Suppose that we process the training examples one at a time.  The SGD update
 for the $i$'th weight matrix is:
 $$
   \W_{ti} = \W_{t{-}1,i} + \eta_t  \x_{ti} \y_{ti}^T 
 $$
  where $\x_{ti}$ is the derivative of the objective function w.r.t. output of the $i$'th weight matrix
  computed at the current sample, and $\y_{ti}$ is the input that the weight matrix acts on.
  These quantities naturally occur in backpropagation.
 
 In our natural gradient method, this is modified as follows:
 $$
   \W_{ti} = \W_{t{-}1,i} + \eta_t \A_{ti}^{-1} \x_{ti} \y_{ti}^T \B_{ti}^{-1},
 $$
 where $\A_{ti}$ and $\B_{ti}$ are factors of the Fisher matrix.  It is easy to show that
 this is equivalent to multiplying the parameter step by the inverse of the Fisher matrix 
 formed from the $\A$ and $\B$ quantities as in Equation~\eqref{eqn:fisher:factorize}.

\subsection{Operation on minibatches}
\label{sec:minibatches}

 Rather than processing training examples one at a time, we process them in minibatches 
 (e.g. 512 at a time).  Instead of vector-valued derivatives $\x_{ti}$ and
 inputs $\y_{ti}$, we now have matrices $\X_{ti}$ and $\Y_{ti}$, each row of which corresponds to one of
 the $\x$ or $\y$ quantities ($t$ is now the index for the minibatch).  The update
 is now as follows:
\begin{equation}
   \W_{ti} = \W_{t{-}1,i} + \eta_t  \X_{ti}^T \Y_{ti}
\end{equation}
 and note that unlike some authors, we don't divide the gradients by the minibatch size-- this
 makes it easier to tune the minibatch size and learning rate independently.
 The update now has the form
\begin{equation}
   \W_{ti} = \W_{t{-}1,i} + \eta_t \bar{\X}_{ti}^T \bar{\Y}_{ti},
\end{equation}
 with the bar indicating modified $\X$ and $\Y$ quantities.  
 In the online version of our natural gradient method, we can write
 these as:
\begin{align}
  \bar{\X}_{ti} &= \X_{ti} \A_{ti}^{-1}  \\
  \bar{\Y}_{ti} &= \Y_{ti} \B_{ti}^{-1},
\end{align}
 but in the simple method, because the $\A$ and $\B$ matrices are estimated from the {\em other}
 elements in the minibatch, we can't write it this way-- it is a separate matrix multiplication for each row of $\X$ and $\Y$--
 but it can still be computed efficiently; see Appendix~\ref{app:simple}.

 In programming terms, we can describe the interface of the core natural-gradient code as follows:
\begin{itemize}
  \item Simple method: Given a minibatch of vectors $\X_{ti}$ with each row being one element of the minibatch,
     estimate the Fisher-matrix factors by holding out each sample, do the multiplication by their inverses,
     and return the modified vectors $\bar{\X}_{ti}$.
  \item Online method: Given a minibatch of vectors $\X_{ti}$ and a previous Fisher-matrix factor $\A_{t-1,i}$,
         compute $\bar{\X}_{ti} = \X_{ti} \A_{t-1,i}^{-1}$ and the updated Fisher-matrix factor $\A_{ti}$.
\end{itemize}
 The interface of the natural gradient code works the same with the $\Y$ and $\B$ quantities, as
 with $\X$ and $\A$.  We call the interface above $2I$ times for each minibatch: twice
 for each weight matrix in the network.

\subsection{Scaling the factors}
\label{sec:scale}

 In both natural gradient methods, we want to prevent the Fisher-matrix multiplication from 
 affecting the overall magnitude of the update very much, compared with the step-sizes in standard SGD. 
 There are several reasons for this:
\begin{itemize}
  \item Early in training, the $\x$ and $\y$ quantities may be very small or zero, leading to huge 
    or infinite inverse-Fisher matrices.
  \item The conventional convergence-proof techniques require that the matrix component of the
      learning rate matrix should have eigenvalues bounded above and below by constants known in advance, 
      which we cannot guarantee if we use an unmodified Fisher matrix.
  \item Empirically, we have found that it is hard to prevent parameter divergence if
       we use the real, un-scaled Fisher matrix.
\end{itemize}
 Our method is to scale the $\bar{\X}_{ti}$ and $\bar{\Y}_{ti}$ quantities so that they have
 the same Frobenius norm as the corresponding inputs $\X_{ti}$ and $\Y_{ti}$.
 We will introduce notation for this in the Appendices.

 This scaling introduces a slight problem for convergence proofs.  The issue is that
 each sample can now affect the value of its own learning-rate matrix (via the
 scalar factor that we use to rescale the matrices).  As we mentioned before,
 it is not permissible in general to use a per-sample learning rate that is a
 function of the sample itself.  However, we don't view this as a practical problem
 because we never use a minibatch size less than 100, so the resulting
 bias is tiny.

 \subsection{Smoothing the Fisher matrix factors with the identity}
\label{sec:smoothing}

 In both versions of NG-SGD, we smooth our estimates of the factors of the Fisher matrix
 by adding a multiple of the identity matrix before inverting them.  In the simple
 method this is necessary because in general the Fisher matrix estimated from the minibatch
 will not be full rank.   In the online method it is not
 strictly necessary because we deal with a factorization of the Fisher matrix
 that already contains a multiple of the unit matrix, but we found that by adding an
 additional multiple of the unit matrix, as for the simple method, we can
 improve the convergence of the SGD training.  In both
 cases the smoothing is of the following form.  If $\S \in \Re^{D \times D} $ is
 a Fisher matrix factor estimated directly from data as the uncentered covariance of the $\x$ or
 $\y$ quantities, then instead of using $\S$ as the Fisher-matrix factor $\A$ or
 $\B$, we use instead $\S + \beta \I$, where
\begin{equation}
  \beta = \textstyle \frac{\alpha}{D} \max( \tr(\S), \epsilon)
\end{equation}
 where $\epsilon = 10^{-20}$ is used to stop the smoothed $\S$ from ever being
 exactly zero.  That is, we smooth the Fisher with the identity matrix scaled by
 $\alpha$ times the average diagonal element of $\S$.  We found in tuning
 experiments that the relatively large value $\alpha = 4$ is suitable under a
 wide range of circumstances, for both the simple and online methods, and even for settings where the
 noise in $\S$ should not be a big problem-- e.g. for large minibatch sizes.  Our
 interpretation is that when $\alpha$ is fairly large, we are using a smaller
 than normal learning rate only in a few directions where the $\x$ or $\y$ quantities 
 have quite high covariance, and a relatively constant learning rate in all the
 remaining directions.

 \section{Comments on the theory}
 \label{sec:prove}

 Although this is not a theoretical paper, we would like to say what we think
 is, and is not, possible to prove about our methods. 

 \subsection{Our factorization of the Fisher matrix}

 If we assume that the distribution of the $\x$ and $\y$ quantities is
 Gaussian and independent (between $\x$ and $\y$ for a single layer, and between layers), then
 it should not be hard to show that the Fisher matrix has the form of~\eqref{eqn:fisher:factorize},
 where the $\A_i$ and $\B_i$ quantities correspond to the uncentered
 covariances of the $\x$ and $\y$ quantities, and that the inverse-Fisher
 has the same form, with the $\A_i^{-1}$ replacing $\A_i$ and $\B_i^{-1}$
 replacing $\B_i$. 

 Of course these conditions won't hold in practical deep learning applications,
 but we do believe that it's a reasonable factorization.  One could try to show
 this experimentally as follows, given a task.  One could make a linear change
 of variables to make our approximated Fisher matrix equal the unit matrix, and
 then try to measure the eigenvalue distribution of the full Fisher matrix in
 the new co-ordinates.  We believe that the eigenvalue distribution of the
 transformed Fisher matrix would probably be much more closerly centered around
 1 than before the change of variables.  Since our motivation for the work
 published here is a practical one, so we have not allocated effort towards this
 type of experiment.

 \subsection{The convergence of our NG-SGD procedure}

 Regarding the convergence of SGD using our factored-Fisher learning rate
 matrices, the most we think is easily provable is that a slightly modified form
 of this method would converge under similar conditions to unmodified SGD.

 The smoothing with constant $\alpha > 0$ can give us a bound on the ratio of
 the largest to smallest eigenvalues of the $\A$ and $\B$ factors; using this
 together with the rescaling of Section~\ref{sec:scale}, we can bound from above
 and below the eigenvalues of the rescaled $\A$ and $\B$ factors.  By
 multiplying these together, we can get lower and upper bounds on the
 eigenvalues of the overall inverse-Fisher matrix that we use as the
 learning-rate matrix $\E_t$.

 It is necessary for the Fisher matrix to be randomly chosen independent of
 the identity of the current sample.  Unfortunately this is not quite true due
 to the rescaling being done at the minibatch level; we mentioned in
 Section~\ref{sec:scale} that this would be a problem for proofs.  As mentioned,
 it would be easy to use the rescaling factor from the previous minibatch; this
 gives us back the independence, but at the cost of no longer having such easy
 bounds on the upper and lower eigenvalues of the rescaled $\A$ and $\B$
 factors.  Alternately, one could keep the algorithm as it is and try to prove
 instead that the parameter value we converge to will not differ very much in some sense from an
 optimum of the true objective function, as the minibatch size gets large.

 \subsection{Online update of a low-rank covariance matrix}

 There might be some interesting things to say about our online natural gradient
 method, described in Appendix~\ref{app:online}, in which estimate the
 uncentered covariance matrices $\A$ and $\B$ in a factored form
 as $\lambda \I + \X \X^T$.  Our online estimation of the covariance matrices involves
 multiplying $\X$ by a weighted combination of (a) the
 observed covariance matrix from the current minibatch, and (b) the previous
 value of our factored approximation to it; it is like a matrix version 
 of the power method~\citep{del1997estimating}.

 Probably the analysis would have to be done initially in the steady state
 (i.e. assuming the parameter vector $\btheta$ is constant).  If in addition we
 assume infinite minibatch size so that the covariance matrix equals its
 expected value, we are confident that we could show that the only stable fixed
 point of our update equations gives us in some suitable sense the closest
 approximation to the covariance; and, with a little more effort, that our
 updates will converge with probability 1 to that best approximation.

 The analysis for finite minibatch size would have to involve different methods.
 Because of the noise and the finite forgetting factor, we would never converge
 to the true value; but it might be possible to define some objective function
 that measures some kind of goodness of approximation, and then say something
 about the convergence of the distribution of that objective function. 

 \subsection{Interaction with other methods}

 We would like to touch on the subject of some other popular modifications of SGD, to explain
 why we do not use them in our experiments.

 One frequently used modification to SGD is momentum~\citep{polyak_momentum}.
 This can be helpful in preventing parameter divergence, as momentum allows SGD
 to use a higher effective learning rate before parameter divergence is
 encountered.  The original reason why none of our experiments involve momentum
 is that we found it quite hard to successfully incorporate momentum into
 multi-threaded parameter updates, needed for the CPU version of our training
 method; this is likely to be the reason why Downpour~\citep{downpour} does not
 use momentum.  We developed other methods to prevent instability-- namely, the
 limit on parameter change per layer per minibatch
 (Appendix~\ref{appendix:weight_limit}); and the natural gradient method itself.

 Another popular modification of SGD is Adagrad~\citep{adagrad}.  This method
 divides the learning rate for each parameter by the standard deviation of the
 sum of gradients for that parameter, averaged over time (from the beginning of
 optimization until the present).  This naturally gives the $1/t$ learning rate
 schedule that is believed from theory to be
 optimal~\citep{kushner2003stochastic}, as well as giving separate learning rates
 for each diagonal element.  There are two reasons why we felt that Adagrad was very
 unlikely to be helpful for large-scale speech recognition.  Firstly, a $1/t$ learning rate has been
 found empirically be inferior to an
 exponentially decaying learning rate~\citep{senior2013empirical}.  Secondly,
 because our $p$-norm nonlinearities~\citep{zhang2014improving} are
 non-saturating we don't believe that our networks are susceptible to the kind of
 pathologies that would make some neurons in a layer require higher learning
 rates than others.  This is also true between different hidden layers, due to special
 properties of the p-norm networks that we use here\footnote{The detailed argument 
  in involves scale invariance of the network output w.r.t. the parameters
  for each layer; an invariance of the learning procedure with respect to scaling up
  the parameters for a layer and scaling up the learning rate at the same time;
  and the notion that parameters in a layer will tend to grow in size due
  to parameter noise, if the learning rate is too high}.
 Essentially, we have reason to believe that as far as some directions requiring
 higher learning rates than others is concerned, all the interesting action for
 our particular type of network is ``off the diagonal''-- that is, it cannot be
 captured by a diagonal matrix.  That is why we have not investigated Adagrad
 and why we smooth our estimates of the factors of the Fisher matrix to the identity
 and not to a diagonal matrix\footnote{Actually, there is another reason for this.  We have
   previously derived an efficient online update of a factored Fisher matrix
   that had a low-rank plus diagonal form (work with Oriol Vinyals, not
   published), and the diagonal term caused the math to become very
   significantly more complicated.}.


\section{Experiments}
\label{sec:experiments}

 We show experiments on a speech recognition setup called Fisher
 English\footnote{Linguistic Data Consortium (LDC) catalog numbers LDC2004S13,
 LDC2005S13, LDC2004T19 and LDC2005T19}, which is English-language
 conversational telephone speech, sampled at 8kHz, and transcribed in a quick
 but relatively low-quality way.  The total amount of training data is 1600
 hours (only including transcribed segments, i.e. not the silent other half of
 the telephone conversation).  We test on a held-out subset of the data,
 about 3.3 hours long, that we defined ourselves.

\subsection{System details and Word Error Rate performance}

\begin{table}[h]
\caption{Word Error Rates (Fisher dev set)} \label{tab:fisher-wers}
\begin{center}
\begin{tabular}{cc}
{\bf Model}\hspace*{0.25in}  & \hspace*{0.25in} {\bf \%WER}  \\ \hline
GMM      &  31.07 \\
DNN1     &  23.66 \\
DNN2     &  23.79  \\
\end{tabular}
\end{center}
\end{table}

Our main results are convergence plots, but to give the reader some idea of the
ultimate results in Word Error Rate, we show some results in
Table~\ref{tab:fisher-wers}.  The Word Error Rates may seem on the high side,
but this is mainly due to the difficulty of the data and the quick transcription
method used on this data.

The GMM system is based on MFCC features, spliced across $\pm 3$ frames and
processed with LDA+MLLT to 40-dimensional features, then adapted with
feature-space MLLR (fMLLR) in both training and test time.
See~\citep{kaldi_paper} for an explanation of these terms and the normal system
build steps.  All these systems used the same phonetic context decision tree
with $7\,880$ context-dependent states; the GMM system had $300\,000$ Gaussians
in total.

The DNN1 system uses speaker adapted features from the GMM system, so it
requires a first pass of GMM decoding and adaptation.  The 40-dimensional
features from GMM1 are spliced across $\pm 4$ frames of context and used as
input to the DNN.  DNN1 is a p-norm DNN~\citep{zhang2014improving} with 5 hidden
layers and p-norm (input, output) dimensions of (5000, 500) respectively,
i.e. the nonlinearity reduces the dimension tenfold.  We use $15\,000$
``sub-classes'' (see Section~\ref{sec:mixture_components} for explanation), and
the number of parameters is 19.3 million.  It is trained for 12 epochs with
learning rate varying from 0.08 to 0.008, trained with 8 parallel jobs with
online natural gradient SGD (NG-SGD).  For both this and the DNN2 system, we
trained with $K = 400\,000$ samples per outer iteration for each machine.

The DNN2 system is trained for our online decoding setup (see
Appendix~\ref{sec:online_decoding}), which is geared towards applications where
reduced latency is important and audio data must be processed strictly in the
order it is received.  The input features are equivalent to unadapted,
un-normalized 40-dimensional log-mel filterbank features, spliced for $\pm 7$
frames, plus a 100-dimensional i-vector representing speaker characteristics,
extracted from only the speaker's audio up to and including the current time.
For the results shown here, we include previous utterances of the same speaker
in the same conversation when computing the i-vector.  Because this system is
intended for real-time decoding on a single CPU, we limit the number of
parameters by using only 4 hidden layers, p-norm (input, output) dimensions of
(350, 3500), and $12\,000$ sub-classes, for a total of 10.4 million parameters.
It was trained using online NG-SGD with 6 parallel jobs for 5 epochs, with the
learning rate decreasing exponentially from 0.01 to 0.001.  All our experiments
below are based on this setup.

Our server hardware is fairly typical: the majority of them are Dell PowerEdge
R720 servers with two Intel Xeon E5-2680v2 CPUs having with 10 cores each,
running at 2.8GHz; and with a single NVidia Tesla K10 GPU card, providing two
GPUs-- each GPU corresponds to a single machine in our notation, and it becomes
incidental that they are co-located.  We also have some similar machines with
K20 GPU cards, and when reporting time taken, we report the slightly more
optimistic figures obtained from running the same jobs on the faster K20 GPUs.

\subsection{Results}

\begin{figure}[th]
\centering
\begin{subfigure}[b]{0.7\columnwidth}
\includegraphics[width=\columnwidth]{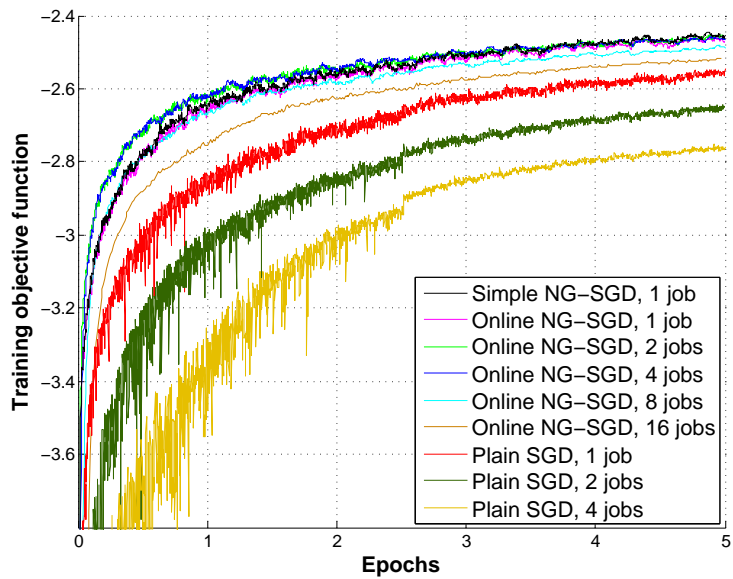}
\caption{Objective function vs. epochs}
\label{fig:epochs}
\end{subfigure}
\begin{subfigure}[b]{0.7\columnwidth}
\includegraphics[width=\columnwidth]{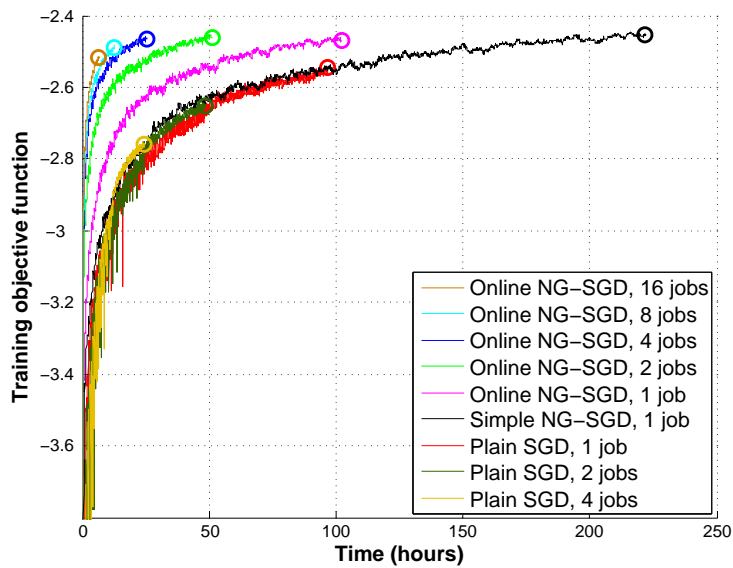}
\caption{Objective function vs. time}
\label{fig:time}
\end{subfigure}
\caption{Convergence of training objective function (log-probability)}
\end{figure}

Our main result is in Figure~\ref{fig:epochs} (best viewed in color), where we
plot the objective function versus amount of training data processed, for our
parallel training method with and without natural gradient, and with 1, 2, 4, 8
and 16 jobs.  In order to keep the effective learning rate
(Section~\ref{sec:overview}) constant, we make the initial/final learning rates
proportional to the number of jobs, with the default learning rates of 0.01 to
0.001 corresponding to the 6-job case.

Our natural gradient method always helps-- the NG-SGD curves are all above the
plain-SGD curves.  Also, when using online natural-gradient, the curves shown in
Figure~\ref{fig:epochs} are close to each other up to about 4 jobs-- i.e. after
processing the same amount of data with different numbers of jobs we get about
the same objective function; however, the 8- and 16-job runs converge a little
slower.  Thus, for small $N$ we are getting a linear speed up in the number $N$
of machines, because the time taken per epoch is proportional to $1/N$.  As $N$
gets larger than around $4$ we need more epochs to get the same improvement, so
the speedup becomes sub-linear.  The plot also shows that the simple and online
natural gradient converge about the same (only tested with one job).  We show
the final Word Error Rates in Table~\ref{tab:ng-wers}; with NG-SGD, they are not 
very sensitive to the number of jobs.

Figure~\ref{fig:time} shows the same plots as Figure~\ref{fig:epochs} but with
time as the x-axis.  This is a simulated clock time, obtained by multiplying the
time taken for each ``outer iteration'' of training, by the number of outer
iterations; the actual clock time depends on queue load.  The time per outer
iteration was 88 seconds for plain SGD, 93 seconds for online NG-SGD, and 208
seconds for plain NG-SGD, all measured on a K20 GPU.  The circles mark the end
of training, after 5 epochs.

\begin{table}[h]
\setlength{\tabcolsep}{3pt}
\begin{center} 
\caption{\small Performance in \%WER} \label{tab:ng-wers}
\end{center}
\begin{center} 
\begin{tabular}{c c c c c c }
\hline
{\small \#jobs N}  & 1 & 2  & 4 & 8 & 16 \\ \hline
 \small Plain SGD  &  23.63  &  23.93  &  24.87  &  &  \\
 \small Simple NG-SGD  &  23.16  &  &   &  &  \\
 \small Online NG-SGD  &  23.19  &  23.00  &  22.84  &  23.12  &  23.35 \\  \hline
\end{tabular}
\end{center}
\end{table}

\section{Conclusions}
\label{sec:conc}

We have described an efficient Natural Gradient version of SGD training
(NG-SGD).  We have shown experimentally that not only does the method improve
the convergence versus plain SGD, it also makes it possible for us to to use a
data parallelization method where we periodically average and redistribute
parameters across multiple SGD runs.  This enables us to train in parallel even
on machines that lack fast interconnections.  Although we only show results from
one setup, we are confident based on past experience that it holds true for
other types of neural network (e.g. ReLU~\cite{mass_relu} or sigmoid
activations) and improves our final results (Word Error Rate) as well as
convergence speed.

We do not have a very good explanation why our parallel training method only
works when using Natural Gradient, except to say that the statements
in~\citep{DBLP:journals/corr/abs-1301-3584} that NG prevents large parameter
steps and is more robust to reorderings of the training set, may be relevant.

\subsubsection*{Acknowledgements}

We would like to thank Karel Vesely, who wrote the original ``nnet1'' neural
network training code upon which the work here is based; Ehsan Variani and Pegah
Ghahremani for their work on CUDA kernels; Hagen Soltau, Oriol Vinyals and
Steven Eliuk for fruitful discussions; and many others, too numerous to mention,
who have contributed to some aspect of the neural net setup and to Kaldi more
generally.

The authors were supported by DARPA BOLT contract No HR0011-12-C-0015, and IARPA
BABEL contract No W911NF-12-C-0015.  We gratefully acknowledge the support of
Cisco Systems, inc. (grant \#574560) and Google, Inc. (Award 2012\_R2\_106,
``Deep Neural Networks for Speech Recognition''), funds which were used to buy
computer equipment and cloud computing time that were used in the development of
these methods.

The U.S. Government is authorized to reproduce and distribute reprints for
Governmental purposes notwithstanding any copyright annotation thereon. The
views and conclusions contained herein are those of the authors and should not
be interpreted as necessarily representing the official policies or
endorsements, either expressed or implied, of DARPA, IARPA, DoD/ARL or the
U.S. Government.

\bibliography{refs}

\newpage







\pagebreak

\appendix

\section{Further details on simple natural gradient method}
\label{app:simple}

 \subsection{Overview of simple natural gradient method}

 In this section we describe the natural gradient method that uses the other elements of the 
 minibatch to estimate the factors of the Fisher matrix.

 As mentioned in Section~\ref{sec:minibatches}, the interface can
 be described as follows.
 Given a matrix $\X$, each row of which represents one element of the minibatch
 (and with a number of columns corresponding to either the row or column dimension
 of one of the weight matrices in the network), do the inverse-Fisher multiplication 
 for each row $\x_i$ of $\X$ and return the modified matrix $\bar{\X}$.

 The core of the inverse-Fisher multiplication is this:
 let $\bar{\x}_i = \F_i^{-1} \x_i$, where $\F_i$ is the Fisher
 matrix estimated from the {\em other} rows of $\X$, i.e. if $N$ is the minibatch
 size, then $\F_i = \frac{1}{N-1} \sum_{j \neq i} \x_j \x_j^T$.
 We extend this basic idea by adding smoothing of $\F_i$ with the identity
 matrix, and by scaling the output $\bar{\X}$ to have the same Frobenius
 norm as the input.

\subsection{Details of method (not considering efficiency)}

 In this section we describe what we compute in our ``simple'' natural gradient
 method, without considering how to compute it efficiently.  As described in
 Section~\ref{sec:smoothing}, we smooth the Fisher matrix with the identity.
 Defining
\begin{equation}
 \beta = \alpha \max( \tr(\X^T \X), \epsilon) / (ND)
\end{equation}
 where our normal settings are $\alpha = 4$ and $\epsilon=10^{-20}$,
 and $N$ and $D$ are the number of rows and columns respectively of $\X$, we define
 smoothed Fisher matrices as follows, to be applied to each row $\x_i$ of $\X$:
\begin{equation}
  \G_i = \left(\beta \I + \frac{1}{N-1} \sum_{j \neq i} \x_j \x_j^T\right)^{-1} 
\end{equation}
 For each row $i$ we will then define
\begin{equation}
 \hat{\x}_i = \G_i^{-1} \x_i
\end{equation}
 and then the result of our computation will be
\begin{equation}
  \bar{\X} = \gamma \hat{\X}
\end{equation}
where the rescaling factor $\gamma$, intended to make sure that $\bar{\X}$ has the same Frobenius norm as the input $\X$,
is defined as
\begin{equation}
  \gamma = \sqrt { \tr(\X^T \X) / \tr(\hat{\X}^T \hat{\X}) } .
\end{equation}
If the denominator of the above is zero, we take $\gamma$ to be one.

We should note that by computing the scalars $\beta$ and $\gamma$ without
``holding out'' the current sample, we are violating the rule that the randomly
sampled learning rate matrix should be independent of the current sample.
However, since we always use a fairly large minibatch size (at least 100) and
these are only scalar quantities, we don't believe the small amount of
``contamination'' that takes place here will significantly bias the training.
In fact, it might not turn out to be very difficult to modify the equations to
properly hold out the current sample for these purposes, but because we don't believe it
would perceptibly affect the results, we haven't gone to the trouble of doing
this.

\subsection{Efficient computation in simple method}

 We now describe how we efficiently compute what we described above.
 Define the smoothed Fisher matrix
\begin{equation}
  \G = \left(\beta\, \I + {\textstyle \frac{1}{N-1}} \X^T \X \right) ,
\end{equation}
 which is like the $\G_i$ quantities but without holding out the current sample.  Next, compute
\begin{equation}
  \Q = \X \G^{-1},
\end{equation}
 where $\Q$ only differs from $\hat{\X}$ by $\x_i$ not being held out from the corresonding $\G_i$.
 There are two equivalent methods to compute $\Q$:
\begin{enumerate}[(i)]
  \item In column space:
$$
    \Q = \X  \left(\beta \I + {\textstyle \frac{1}{(N-1)}} \X^T \X \right)^{-1}
$$
  \item  In row space:
$$
    \Q = \left(\beta \I + {\textstyle \frac{1}{(N-1)}} \X \X^T \right)^{-1} \X 
$$
\end{enumerate}
 We derived the rather surprising row-space version of the formulation by expanding the inverted
 expression on the right of the column-space expression using the Morrison-Woodbury formula, and
 simplifying the resulting expression.

For efficiency, we choose method (i) above if the minibatch size is greater than
the dimension ($N > D$), and method (ii) otherwise.  Our formula below for
$\hat{\X}$ is derived by expressing each $\G_i^{-1}$ as a rank-one correction to $\G^{-1}$,
and computing the corresponding correction by which each row $\hat{\x}_i$ differs from
the corresponding row $\q_i$ of $\Q$.  It turns out that the correction is in the same 
direction as $\q_i$ itself, so $\hat{\x}_i$ just becomes a scalar multiple of $\q_i$.
Defining, for each row-index $i$,
\begin{equation}
  a_i = \x_i^T \q_i ,
\end{equation}
and defining the scalar factor
\begin{equation}
  b_i = 1 + a_i / (N {-} 1 {-} a_i),
\end{equation}
then we can use the following efficient formula: for each row $\hat{\x}_i$ of $\hat{\X}$,
\begin{equation}
  \hat{\x}_i = b_i \q_i .
\end{equation}
 We then get the output $\bar{\X}$ by scaling $\hat{\X}$ by $\gamma$, as described above.
 
 When working on CPUs with small minibatch sizes (e.g. $N = 128$) and large
 hidden-layer dimensions (e.g. $D = 1000$), the computation above is very
 efficient, and does not comprise more than about 20\% of the time of the
 overall backprop computation.  However, when using GPUs with larger minibatch
 sizes (e.g. $N = 512$) it can take the majority of the time.  Even though it
 typically takes considerably less than half of the total floating point
 operations of the overall computation, it contains a matrix inversion, and
 matrix inversions are not very easy to compute on a GPU.  Our ``online'' method
 which we will describe below is designed to solve this efficiency problem.

\section{Further details on online natural gradient method}

 \label{app:online} 

 \subsection{Overview of online natural gradient method}

 The interface of the online natural-gradient method is essentially the same as
 the simple method: the user provides a matrix $\X$, and we return a matrix
 $\bar{\X}$ that's been multiplied by the inverse-Fisher and then rescaled to
 have the same Frobenius norm as $\X$.  Again, each row of $\X$ corresponds to
 an element of the minibatch and the column dimension corresponds to the row or
 column dimension of one of the weight matrices.  A difference from the simple method
 is that the online method is ``stateful'', because we maintain a running estimate of the Fisher matrix.  Each
 time we process a minibatch, we use the Fisher matrix estimated from the
 previous minibatches; and we then update that estimate using the current
 minibatch.  For a single neural net, the number of separate copies of this
 ``state'' that we need to maintain corresponds to twice the number of trainable
 weight matrices in the neural net: one for each of the $\A_i$ and $\B_i$
 quantities in Equation~\eqref{eqn:fisher:factorize}.

 Let the input be $\X \in \Re^{N \times D}$, where $N$ is the minibatch size
 (e.g. 512) and $D$ is the row or column size of the weight matrix
 we're updating (e.g. 2000).  We introduce a user-specified parameter $R < D$ which is the
 rank of the non-identity part of the Fisher matrix. Let
 the subscript $t = 0, 1,\ldots$ correspond to the minibatch.  Define
\begin{equation}
 \label{eqn:low:rank}
   \F_t  \eqdef \R_t^T \D_t \R_t + \rho_t \I
\end{equation}
 where $\R_t \in \Re^{R \times D}$, $\D_t \in \Re^{R \times R}$ and $\rho_t > 0$ will be 
 estimated online from data; $\R_t$ has orthonormal rows and $\D_t$ is diagonal and nonnegative.
 We'll estimate these quantities online from the data with the aim being that $\F_t$ should be
 a good estimate of the covariance of the rows of the $\X_t$ quantities.

 We define $\G_t$ to be a kind of ``smoothed'' version of $\F_t$ where we add in
 more of the unit matrix, controlled by the $\alpha$ parameter we've previously discussed
 (normally $\alpha = 4$):
\begin{equation}
      \G_t  \eqdef  \F_t +  \frac{\alpha \tr(\F_t)}{D} \I  .
\end{equation}
 and then the output will be: 
\begin{equation}
   \bar{\X}_t = \gamma_t \X_t \G_t^{-1}
\end{equation} 
 where $\gamma_t$ is computed so as to ensure that the Frobenius norm of $\bar{\X}_t$ equals that of $\X_t$: 
\begin{equation}
  \gamma_t = \sqrt{ \tr(\X_t \X_t^T) / \tr( \X_t \G_t^{-1} \G_t^{-1} \X_t^T ) } ,  \label{eqn:gammat:old}
\end{equation}
 or $\gamma_t = 1$ if the denominator of the above equation is 0.

\subsection{Updating our low-rank approximation to the variance}

 Next we discuss the method we use to estimate our low-rank approximation $\F_t$ of
 the uncentered covariance of the rows of the inputs $\X_t$.  Define
\begin{equation}
  \S_t  \eqdef {\textstyle \frac{1}{N}} \X_t^T \X_t
\end{equation}
 as the uncentered covariance of the rows of $\X_t$.  We introduce a
 user-specified ``forgetting factor'' $0 < \eta < 1$ (we describe how this is
 set in Section~\ref{sec:b:configuration}), and we define
\begin{equation}
  \T_t  \eqdef \eta \S_t + (1-\eta) \F_t .
\end{equation} 
 We will try to set $\F_{t{+}1}$ to be a good low-rank approximation to $\T_t$.
 The obvious way would be to make $\D_{t{+}1}$ correspond to the top eigenvalues
 of $\T_t$ and $\R_{t{+}1}$ to the corresponding eigenvectors, but this would be
 too slow.  Instead we use a method inspired by the power method for finding the
 top eigenvalue of a matrix.  On each iteration we compute
\begin{equation}
\label{eqn:yt}
  \Y_t  \eqdef  \R_t \T_t,
\end{equation}
 with $\Y_t \in \Re^{R \times D}$.  It is useful to think of $\Y_t$ as containing each
 eigenvector scaled by its corresponding eigenvalue in $\T_t$ (of course, this is true
 in a precise sense only at convergence).  Our update uses symmetric eigenvalue decomposition of $\Y_t \Y_t^T$ to
 find these scaling factors (they are actually the square roots of the eigenvalues of $\Y_t \Y_t^T$),
 puts them on the diagonal of $\D_{t{+}1}$, and puts the corresponding
 eigenvectors in the rows of $\R_{t{+}1}$.  We then have to work out the correct amount of the unit-matrix
 to add into our factorization of the covariance matrix (i.e. set $\rho_{t{+}1}$) and subtract that amount 
 from the diagonals of $\D_{t{+}1}$.  We will give equations for this below.

 Observant readers might have noted that it would seem more straightforward do
 do a Singular Value Decomposition (SVD) on $\Y_t$ instead of a symmetric
 eigenvalue decomposition on $\Y_t \Y_t^T$.  We do it this way for speed.

 The details of our update are as follows: 
\begin{equation}
  \Z_t \eqdef \Y_t \Y_t^T,  \label{eqn:zt:def}
\end{equation}
 so $\Z_t \in \Re^{R\times R}$.  Then do the symmetric eigenvalue decomposition
\begin{equation}
    \Z_t = \U_t \C_t \U_t^T,     \label{eqn:zt:eig}
\end{equation}
 with $\U$ orthogonal and $\C_t$ diagonal.  The diagonal elements of $\C_t$ are positive;
 we can prove this using $\rho_t > 0$ (which makes $\T_t$ positive definite) and using the
 fact that $\R_t$ has full row rank.  We define $\R_{t{+}1}$ as:
\begin{equation}
\label{eqn:rt1:def}
   \R_{t{+}1} \eqdef \C_t^{-0.5} \U_t^T \Y_t
\end{equation}
 If we expand out $\R_{t{+}1} \R_{t{+}1}^T$ using~\eqref{eqn:rt1:def}, it is easy to see that it
 reduces to the identity, hence $\R_{t{+}1}$ has orthonormal rows.  In order to make sure
 that $\F_{t{+}1}$ has the desired covariance in the directions corresponding to the rows of $\R_{t{+}1}$,
 we will set
\begin{equation}
  \D_{t{+}1}  \eqdef \C_t^{0.5} - \rho_{t{+}1} \I,
\end{equation}
 but note that at this point, $\rho_{t{+}1}$ is still unknown.  When we say the ``desired covariance'',
 we are ensuring that for each dimension $\r$ corresponding to a row of $\R_{t{+}1}$, the value of
 the inner product $\r^T \F_{t{+}1} \r$ equals that of $\r^T \T_t \r$, but this is only precisely true at
 convergence. 

 We choose $\rho_{t{+}1}$ in order to ensure that $\tr(\F_{t{+}1}) = tr(\T_t)$.  This
 value can be worked out as:
\begin{equation}
\label{eqn:rhodash}
  \rho'_{t{+}1} = {\textstyle \frac{1}{D{-}R}}  \left( \eta \tr(\S_t) {+} (1{-}\eta)(D \rho_t {+} \tr(\D_t)) {-} \tr(\C_t^{0.5}) \right)
\end{equation}
 We then let
\begin{equation}
    \rho_{t{+}1} = \max(\epsilon, \rho'_{t{+}1})
\end{equation}
 for $\epsilon = 10^{-10}$; this is is to ensure that if we get a sequence of zero inputs,
 $\rho_t$ will not become exactly zero in its machine representation.

 \subsection{Efficient computation}
\label{sec:b:efficient}

 The previous section described what we are computing in the online natural
 gradient method; here we describe how to compute it efficiently.  The essential
 idea here is to reduce the multiplication by $\G^{-1}$ to two multiplications
 by a ``fat'' matrix (of dimension $R \times D$).  Since typically $R$ is
 much smaller than $D$, this is quite efficient.  We also address how to
 efficiently keep these matrices updated, at the level of optimizing the matrix
 expressions.  This section is mostly derivation, and will likely only be of
 interest to someone who is considering implementing this method.  In
 Section~\ref{sec:online:summary} below, we will summarize the algorithm we
 derive here.

 We can write $\G_t$ as:
\begin{align}
       \G_t &\eqdef    \F_t +  {\textstyle \frac{\alpha \tr(\F_t)}{D}} \I  \label{eqn:gt} \\
            &=         \R_t^T \D_t \R_t + \beta_t I  \\
\shortintertext{where}
    \beta_t &\eqdef    \rho_t + {\textstyle \frac{\alpha}{D}} \tr(\F_t)               \\
            &=         \rho_t(1+\alpha) + {\textstyle \frac{\alpha}{D}} \tr(\D_t)    \label{eqn:beta2}
\end{align}
 Define
\begin{equation}
 \label{eqn:hatxt}
   \hat{\X}_t \eqdef \beta_t \X_t \G_t^{-1} ,
\end{equation}
 where the factor of $\beta_t$ is inserted arbitrarily to simplify the update equations; a scalar
 factor on $\hat{\X}$ doesn't matter because we will later rescale it to have the same norm
 as $\X$.  The output of this whole process is
\begin{align}
  \bar{\X}_t  &\eqdef \gamma_t \hat{\X}_t \mbox{, where}        \\
  \gamma_t    &\eqdef \sqrt{ \tr(\X_t \X_t^T)  / \tr(\hat{\X}_t^T \hat{\X}_t) } , \label{eqn:gammat}
\end{align}
where, in the expression for $\gamma_t$, if the denominator is zero we take $\gamma_t = 1$.
Note: $\gamma_t$ is not the same as in~\eqref{eqn:gammat:old} because of the arbitrary factor of 
$\beta_t$, so consider~\eqref{eqn:gammat:old} to be superseded by~\eqref{eqn:gammat}.
To efficiently compute~\eqref{eqn:hatxt}, we apply the Woodbury matrix identity to~\eqref{eqn:gt}, giving 
us
\begin{align}
  \G_t^{-1} &= \frac{1}{\beta_t} \left( \I - \R_t^T \E_t \R_t   \right)     \\
\shortintertext{where}
   \E_t    &\eqdef \frac{1}{\beta_t} \left( \D_t^{-1} + \frac{1}{\beta_t} \I \right)^{-1}  \\
 \shortintertext{with elements}
   e_{tii}  & =  \frac{1}{\beta_t/d_{tii} + 1 }    \label{eqn:etii}
\end{align}
In order to reduce the number of matrix multiplies, it is useful to break the expression 
$\R_t^T \E_t \R_t$ into two equal parts, so we define
\begin{equation}
   \W_t \eqdef \E_t^{0.5} \R_t ,  \label{eqn:wt:def}
\end{equation}
and we will never store $\R_t$; instead, we will work with $\W_t$ and the small diagonal
factors $\D_t$ and $\E_t$.  We can now write the following, which is where most of our computation will take place:
\begin{equation}
   \hat{\X}_t  =  \X_t - \X_t \W_t^T \W_t  \label{eqn:hatxt:compute}
\end{equation}
You may recall the symmetric matrix $\Z_t \in \Re^{R \times R}$ defined in~\eqref{eqn:zt:def}, which
is involved in the update of our factorization.  The following expressions are going to be useful
when computing it, and the first of them appears as a sub-expression of~\eqref{eqn:hatxt:compute}.
For convenience we state the dimensions of these quantities below:
\begin{align}
     \H_t &\eqdef  \X_t \W_t^T   \in \Re^{N \times R}  \label{eqn:ht} \\
     \J_t &\eqdef \H_t^T \X_t    \in \Re^{R \times D} \\
          & =     \W_t \X_t^T \X_t                   \\ 
     \K_t &\eqdef \J_t \J_t^T    \in \Re^{R \times R} \mbox{(symmetric)} \\
     \L_t &\eqdef \H_t^T \H_t    \in \Re^{R \times R} \mbox{(symmetric)} \\
          & =     \W_t \X_t^T \X_t \W_t^T   \\
          & =     \J_t \W_t^T              
\end{align}
After we have $\H_t$, we can compute $\hat{\X}_t$ using a single matrix multiply as:
\begin{equation}
   \hat{\X}_t  =  \X_t - \H_t \W_t .  \label{eqn:hatxt:compute:2}
\end{equation}
We can expand $\Y_t = \R_t \T_t$, defined in~\eqref{eqn:yt}, into quantities that will be computed, as:
\begin{align}
    \Y_t &=  {\textstyle \frac{\eta}{N}} \R_t \X_t^T \X_t + (1{-}\eta) (\D_t + \rho_t \I) \R_t   \\
         &=  {\textstyle \frac{\eta}{N}} \E_t^{-0.5} \J_t + (1{-}\eta) (\D_t + \rho_t \I) \E_t^{-0.5} \W_t  \label{eqn:yt:compute}
\end{align}
Using~\eqref{eqn:yt:compute} we can expand $\Z_t = \Y_t \Y_t^T$, as:
\begin{multline}
    \Z_t   = {\textstyle \frac{\eta^2}{N^2}} \E_t^{-0.5} \J_t \J_t^T \E_t^{-0.5}  +   (1{-}\eta)^2 \left(\D_t + \rho_t \I\right)^2  \\
              + {\textstyle \frac{\eta(1{-}\eta)}{N}} \E_t^{-0.5} \J_t \W_t^T \E_t^{-0.5} (\D_t + \rho_t \I)        \\
              + {\textstyle \frac{\eta(1{-}\eta)}{N}} (\D_t + \rho_t \I) \E_t^{-0.5} \W_t \J_t^T \E_t^{-0.5}
\end{multline}
and we can substitute some of the sub-expressions we defined above into this, to give:
\begin{multline}
    \Z_t   = {\textstyle \frac{\eta^2}{N^2}}    \E_t^{-0.5} \K_t \E_t^{-0.5}  +   (1{-}\eta)^2 \left(\D_t + \rho_t \I\right)^2  \\
              + {\textstyle \frac{\eta(1{-}\eta)}{N}}  \E_t^{-0.5} \L_t \E_t^{-0.5} (\D_t + \rho_t \I)        \\
              + {\textstyle \frac{\eta(1{-}\eta)}{N}} (\D_t + \rho_t \I) \E_t^{-0.5} \L_t \E_t^{-0.5}  \label{eqn:zt:compute}
\end{multline}
Our strategy will be to compute the symmetric quantities $\L_t$ and $\K_t$ on the GPU, and transfer
them to the CPU where we can then compute $\Z_t$ using the expression above -- this can be done in $O(R^2)$ -- 
and then do the symmetric eigenvalue decomposition as in~\eqref{eqn:zt:eig}, on the CPU.  We repeat the
equation here for convenience:
\begin{equation}
 \label{eqn:zt:eig:repeat}
    \Z_t = \U_t \C_t \U_t^T . 
\end{equation}
Here, $\U_t$ will be orthogonal, and mathematically, no element of the
diagonal matrix $\C_t$ can be less than $(1{-}\eta)^2 \rho_t^2$, so we floor its diagonal
to that value to prevent problems later if, due to roundoff, any element 
is smaller than that.

Below, we'll say how we efficiently compute $\tr(\X \X^T)$ and $\tr(\hat{\X} \hat{\X}^T)$; for
now, just assume those quantities have been computed.

We compute $\rho_{t{+}1}$ as follows, expanding $\S_t$ in~\eqref{eqn:rhodash}:
\begin{multline}
  \rho'_{t{+}1} =  \frac{1}{D{-}R} \bigg( \frac{\eta}{N} \tr(\X \X^T) +   \\
                   (1{-}\eta)(D \rho_t {+} \tr(\D_t)) - \tr(\C_t^{0.5}) \bigg) . \label{eqn:rhodash2}
\end{multline}
We can now compute $\D_{t{+}1}$ and $\rho_{t{+}1}$; we floor both to $\epsilon$ to ensure they
never go to exactly zero which could cause problems for our algorithm.
\begin{align}
     \D_{t{+}1} &= \max( \C_t^{0.5} - \rho'_{t{+}1} \I, \epsilon \I )    \label{eqn:dt1} \\
  \rho_{t{+}1}  &=  \max(\epsilon, \rho'_{t{+}1})  \label{eqn:rhot1}
\end{align}
for a small constant $\epsilon = 10^{-10}$ (the first $\max$ is taken per element).
We can now compute the scalar $\beta_{t{+}1}$ and the diagonal matrix $\E_{t{+}1}$ (we show the formula for its diagonal elements):
\begin{align}
    \beta_{t{+}1} &= \rho_{t{+}1} (1{+}\alpha) + \frac{\alpha}{D} \tr(\D_{t{+}1})   \\
        e_{tii} &= \frac{1}{ \beta_{t{+}1}/d_{t{+}1,ii} + 1 }
\end{align}
We never construct $\R_{t{+}1}$ in memory, but instead we directly compute
$\W_{t{+}1}$.  We can factor it as follows:
\begin{align}
  \W_{t{+}1} &\eqdef \E_{t{+}1}^{0.5} \R_{t{+}1}   \\
                &= \E_{t{+}1}^{0.5} \C_t^{-0.5} \U_t^T \Y_t  \\
                &= \E_{t{+}1}^{0.5} \C_t^{-0.5} \U_t^T \left( {\textstyle \frac{\eta}{N}} \E_t^{-0.5} \J_t  + (1{-}\eta) (\D_t {+} \rho_t \I) \R_t \right) \\
                &= \A_t \B_t    \label{eqn:wt1} \\
\shortintertext{where}
    \A_t      &\eqdef {\textstyle \frac{\eta}{N}} \E_{t{+}1}^{0.5} \C_t^{-0.5} \U_t^T \E_t^{-0.5}  \\
    \B_t      &\eqdef \J_t + {\textstyle \frac{N(1{-}\eta)}{\eta}} \left(\D_t + \rho_t \I \right) \W_t ,
\end{align}
and note that while it might seem like a factor of $\E_t^{-0.5}$ is missing from the second term in $\B_t$,
in fact we use the fact that it commutes with $\left(\D_t + \rho_t \I \right)$ to move it to the left, into $\A_t$.
If we're using a GPU, $\A_t$ will be computed in time $O(R^2)$ on the CPU and transferred to the
GPU; we then compute $\B_t$ on the GPU efficiently by scaling the rows of $\W_t$ and adding $\J_t$;
then we multiply $\A_t$ and $\B_t$ on the GPU.

\subsubsection{Maintaining orthogonality}
\label{sec:orthogonality}

We have noticed that the invariance $\R_t \R_t^T = \I$ can sometimes be lost due to
roundoff.  A proper analysis of roundoff in our algorithm is not something we
have time to do, but we will describe how we detect and fix this problem in practice.  For
speed, we only do the following operations if the diagonal matrix $\C_t$,
has condition number greater than $10^{6}$, or if any elements
were floored as mentioned just after~\eqref{eqn:zt:eig:repeat}.
Note: all the computations we describe in this paper were done in single precision.

We compute the symmetric matrix
\begin{align}
   \O_t &\eqdef  \R_t \R_t^T  \\
         & =   \E_t^{-0.5} \left(\W_t \W_t^T\right) \E_t^{-0.5} ,
\end{align}
where the part in parentheses is computed on the GPU and transferred to the CPU.
If no element of $\O_t$ differs by more than $10^{-3}$ from the corresponding element
of the unit matrix, we consider that $\R_t$ is sufficiently orthogonal and we do nothing
more.  Otherwise, we do a Cholesky decomposition $\O_t = \C \C^T$, compute
the reorthogonalizing factor $\M = \E_t^{0.5} \C^{-1} \E_t^{-0.5}$ on the CPU and copy to the GPU,
and do $\W_{t{+}1} \leftarrow \M \W_{t{+}1}$ to reorthogonalize.  Re-orthogonalization
happens extremely rarely, and usually only if something bad has already happened such as
parameter divergence.

\subsubsection{Initialization}
\label{sec:initialize}

In our implementation we don't bother dumping the ``state' of the computation to
disk so each new process reinitializes them for the first minibatch it
processes.  We initialize them so as to most closely approximate the covariance
of the first minibatch of features.  This is done by taking
\begin{equation}
  \S_0  \eqdef {\textstyle \frac{1}{N}} \X_0^T \X_0
\end{equation}
and finding the top $R$ eigenvalues and eigenvectors; the rows of $\R_0$ contain
the top eigenvectors.  Let $\lambda_i$ be the corresponding eigenvalues, for $1 \leq i \leq R$,
and we set
\begin{equation}
  \rho_0 = \max\left( \frac{ \tr(\S_0) - \sum_{i=1}^R \lambda_i }{ D - R }, \epsilon \right)
\end{equation}
for $\epsilon = 10^{-10}$, and for $1 \leq i \leq R$, 
we let $d_{0ii} \leftarrow \max(\epsilon, \lambda_i - \rho_0)$.

\subsubsection{Computing matrix traces}

We mentioned above that we have a fast way of computing the quantities $\tr(\X
\X^T)$ and $\tr(\hat{\X} \hat{\X}^T)$.  These are needed to compute $\gamma_t$
using~\eqref{eqn:gammat}, and to compute $\rho'_{t{+}1}$
using~\eqref{eqn:rhodash2}.  We compute these as a side effect of the fact that
we need, for each row $\bar{\x}_{ti}$ of the output, its squared norm
$\bar{\x}_{ti}^T \bar{\x}_{ti}$.  This will be required to enforce the
``maximum parameter change'' per minibatch, as described in
Section~\ref{sec:max_param_change}.  Suppose we've already computed $\hat{\X}_t$
using~\eqref{eqn:hatxt:compute:2}.  We compute
the inner products for all rows $1 \leq i \leq N$ of $\hat{\X}_t$ as
\begin{equation}
  p_i =  \hat{\x}_{ti}^T \hat{\x}_{ti}^T ,  \label{eqn:pi}
\end{equation}
using a single GPU kernel invocation.   If we are updating the parameters of the Fisher-matrix factorization,
then we can most efficiently obtain our desired traces as follows:
\begin{align}
  \tr(\hat{\X} \hat{\X}^T)  &= {\textstyle \sum_i} p_i       \\
    \tr(\X \X^T)            &= \tr(\hat{\X} \hat{\X}^T) - \tr(\L_t \E_t) + 2 \tr(\L_t) .  \label{eqn:trxxt}
\end{align}
The expression for $\tr(\X \X^T)$ was obtained by expanding $\tr(\hat{\X} \hat{\X}^T)$
using~\eqref{eqn:hatxt:compute:2}, moving $\tr(\X \X^T)$ to the left, and recognizing 
sub-expressions that we have already computed.
In case we are not updating the parameters of the Fisher-matrix factorization, we have no
other need for $\L_t$ so it will be more efficient to compute $\tr(\X \X^T)$ directly; this
can of course be done in $O(N D)$ operations and does not require a matrix multiply.
Once we have the scaling factor $\gamma_t$ we can scale the $p_i$ quantities by its
square, and they will equal the quantities $\bar{\x}_{ti}^T \bar{\x}_{ti}$ that we'll need
for enforcing the maximum parameter change.

\subsubsection{Multithreading and other issues}

Most of what we have written above is geared towards operation using a GPU, but
we also support operation with CPUs, where our SGD implementation is
multithreaded.  In this case, we have to consider the interaction with
multithreaded code because of the ``stateful'' nature of the computation.  We
wanted to avoid a bottleneck where different threads wait to update the
parameters sequentially.  Our solution is that before doing the part of the
computation where we update the parameters, we try to get a lock, and if this
fails, we simply apply the fixed inverse-Fisher matrix but don't update the
Fisher-matrix parameters $\R_t$ and so on.  Since the model parameters don't
move very fast, we don't expect that this will make any noticeable difference
to the SGD convergence, and we have seen no evidence that it does.

\subsection{Typical configuration}
\label{sec:b:configuration}

The most important user-specified parameters for our algorithm are the rank $R$
and the constant $\alpha$ that controls smoothing with the unit matrix.  The
value $\alpha = 4$ seems to work well over a wide variety of conditions, so we
normally leave it at that value.  The rank $R$ should generally increase with
the dimension of the vectors we are multiplying.  Our experiments here are with
``p-norm'' networks where the nonlinearity is dimension reducing, like
maxout~\citep{gdfl_maxout}, typically reducing the dimension from something like
3000 to 300.  So a typical parameter matrix will increase the dimension from
something like 301 to 3000 (it's 301 instead of 300 because of the bias term).
Our normal rule for ranks is to use $R = 20$ on the input side of each matrix
and $R = 80$ on the output side.  Part of the way we originally tuned this is to
look at the diagonal matrices $\E_t$.  These matrices have diagonal values $0 <
e_{tii} < 1$, sorted on $i$ from greatest to least, and $1 - e_{tii}$ can be
interpreted as the amount by which the input is scaled in a certain direction in
the space.  A value of $e_{tii}$ close to 1 means we are strongly scaling down
the input, and a value close to 0 means we are leaving it unchanged.  If the
last $e_{tii}$ has a value of, say, 0.1, then reducing $R$ by one will be like
taking a scaling factor of 0.9 applied to a gradient, and setting to 1 instead;
this seems unlikely to make any difference to the SGD, as it's like changing the
learning rate in some direction from 0.9 to 1.  Our final $e_{tii}$ values are
normally in the range 0.05 to 0.2.

Another configurable constant is the ``forgetting factor'' $0 < \eta < 1$: the
closer $\eta$ is to 1, the more rapidly we track changes in the Fisher matrix
due to changes in parameters, but the more noise we will have in our estimates.
Because we don't want to have to tune $\eta$ when we change the minibatch size,
we set it as follows.  The user specifies a parameter $S$ (interpreted as
an approximate number of samples to include in our estimate of the Fisher
matrix), and we set
\begin{equation}
 \eta = 1 - \exp(-N / S),  \label{eqn:eta:ns}
\end{equation}
where $N$ is the minibatch size.  We normally set $S = 2000$; we have no reason
to believe that this is a very important parameter.

In order to increase the speed of the algorithm, we normally configure it so
that we only actually update the parameters of the Fisher matrix every 4 minibatches,
except on the first 10 minibatches in a process, when we always update them.

\subsection{Summary of the online natural gradient method}
\label{sec:online:summary}

Here we summarize the online natural-gradient SGD method-- that is, we summarize
the core part of the algorithm that takes a matrix $\X \in \Re^{N \times D}$,
and outputs a matrix $\bar{\X} \in \Re^{N \times D}$.
To understand how this fits into the bigger picture of back-propagation and SGD,
see Section~\ref{sec:natural:gradient}.

For this summary we will ignore issues of multithreading.  Our explanation
here is just for one instance of the algorithm, corresponding to the row or column
dimension of one of the weight matrices; if there are $I$ weight matrices,
there are $2I$ separate copies of the variables we describe here.

Typical configuration variables are as follows: $\alpha = 4$, $S = 2000$ (this
will determine $\eta$), rank $R = 20$ (or 80), $\epsilon = 10^{-10}$; and let's
define a variable $J = 4$ that dictates the period with which we update the
Fisher-matrix factors.  Minibatch size $N$ is normally 128 (on CPU), or 512 (on GPU).

On $t = 0$, before running the steps below we have to initialize the parameters
as described in Section~\ref{sec:initialize}.  Note: while in Section~\ref{sec:initialize}
we describe how to set $\rho_0$, $\R_0$ and $\D_0$, the variables which we actually store 
are $\rho_0$, $\D_0$, and $\W_0$; to compute $\W_0$ we need 
Equations~\eqref{eqn:beta2},~\eqref{eqn:etii} and~\eqref{eqn:wt:def}.

We have an input $\X \in \Re^{N \times D}$, and despite the
notation, we do not require that $N$ be the same for all $t$-- sometimes the
last minibatch we process has a smaller than normal size.

If $t < 10$ or $J$ divides $t$ exactly, then we will be updating the factored
Fisher matrix; otherwise we just apply it and don't update.  There are two
slightly versions of the algorithm, depending whether we will be updating the
Fisher matrix.

In either case, we first compute $\eta$ from $N$ and $S$ using~\eqref{eqn:eta:ns},
and then compute 
\begin{equation}
  \H_t =  \X_t \W_t^T  .
\end{equation}
From here the two cases begin to differ.

\paragraph{Without updating the Fisher matrix.}

If we won't be updating the Fisher matrix, then it's simpler.
The input is $\X_t$.  We first compute $\tr(\X_t^T \X)$.
Then we compute
\begin{equation}
  \hat{\X}_t = \X_t - \H_t \W_t,
\end{equation}
overwriting the input $\X_t$.
Next, for each $1 \leq i \leq N$ we compute the row-products $p_i$ using~\eqref{eqn:pi},
and compute $\tr(\hat{\X}^T \hat{\X})$ as the sum of $p_i$.  Now we can compute $\gamma_t$
using~\eqref{eqn:gammat}.  Next we scale $\hat{\X}_t$ by $\gamma_t$ to produce $\bar{\X}_t$.
We also output for each $i$ the quantity $\gamma_t^2 p_i = \bar{\x}_{ti}^T \bar{\x}_{ti}$, which
is needed to enforce the ``maximum parameter change per minibatch'' constraint.

\paragraph{With updating the Fisher matrix.}

If we're updating the Fisher matrix, which we usually do every four steps, there are
some more operations to do.   First we compute
\begin{equation}
  \J_t = \H_t^T \X_t \in \Re^{R \times D} .
\end{equation}
Next we want to compute $\L_t$ and $\K_t$.  We actually have two separate strategies for this.
If $N > D$ (the minibatch size exceeds the vector dimension), we do:
\begin{align}
  \L_t &= \W_t \J_t^T \in \Re^{R \times R} \\
  \K_t &= \J_t \J_t^T \in \Re^{R \times R}
\end{align}
and in our implementation we combine these into one matrix operation by placing
$\L$ and $\K$, and $\W$ and $\J$, next to each other in memory.  Otherwise, we compute
$\K_t$ as above but $\L_t$ using:
\begin{equation}
  \L_t  = \H_t^T \H_t  \in \Re^{R \times R}.
\end{equation}
At this point, if we're using a GPU, we transfer the symmetric
matrices $\K_t$ and $\L_t$ to the CPU.
We now compute some small derived quantities on the CPU: $\beta_t$
using~\eqref{eqn:beta2} and $\E_t$ using~\eqref{eqn:etii}, as well as
$\E_t^{0.5}$ and $\E_t^{-0.5}$; $\E_t$ is diagonal so this is not hard.
At this point we compute the symmetric $R\times R$ matrix $\Z_t$ using~\eqref{eqn:zt:compute};
the expression looks scary but it can be computed in $O(R^2)$ time.

We do the symmetric eigenvalue decomposition as in~\eqref{eqn:zt:eig:repeat},
on the CPU, to get the orthogonal matrix $\U_t$ and the diagonal matrix $\C_t$,
and we floor the diagonal elements of $\C_t$ to $(1{-}\eta)^2 \rho_t^2$. 

Next we compute
\begin{equation}
  \hat{\X}_t = \X_t - \H_t \W_t,
\end{equation}
then compute the row-products $p_i$ using~\eqref{eqn:pi}, compute
$\tr(\hat{\X}_t^T \hat{\X}) = \sum_i p_i$, and
compute $\tr(\X_t^T \X)$ using~\eqref{eqn:trxxt}.  We can now obtain
the scaling factor $\gamma_t$ using~\eqref{eqn:gammat}, and use it to compute
the main output $\bar{\X}_t = \gamma_t \hat{\X}_t$ and the per-row
inner products of the output which equal $\gamma_t^2 p_i$ (although in
our implementation, to save time we actually output $\gamma_t$ and let the 
user do the scaling later on).

We next compute $\rho'_{t{+}1}$ using~\eqref{eqn:rhodash2}, $\D_{t{+}1}$ using~\eqref{eqn:dt1}
and $\rho_{t{+}1}$ using~\eqref{eqn:rhot1}.
$\W_{t{+}1}$ is computed using a matrix multiply on the GPU as in~\eqref{eqn:wt1},
after working out the factors $\A_t$ and $\B_t$.

At this point, if we had floored any diagonal elements of $\C_t$ above or
if its condition number after flooring exceeds $10^6$, we do the orthogonality check
and possible reorthogonalization that we described in Section~\ref{sec:orthogonality}
above.

\section{Other aspects of our DNN implementation}
\label{appendix:other_aspects}

 Here we describe some aspects of our neural net training implementation that are of less direct
 relevance to our parallel training and natural gradient methods, so were not included in the main text
 of the paper.

 In Section~\ref{appendix:cpu:gpu} we discuss the CPU-based and GPU-based versions of our
 SGD implementation and how they differ; in Section~\ref{appendix:data:organize} we discuss
 how we randomize the training examples and store them on disk.
 In Section~\ref{appendix:weight_limit} we explain how we enforce a maximum
 parameter-change per minibatch; in~\ref{sec:model_averaging} we explain our
 generalized model-averaging procedure; in~\ref{sec:mixture_components} we
 explain how we use ``mixture components'' (a.k.a. sub-classes) for DNNs;
 in~\ref{sec:input_data_normalize} we introduce our method of input data
 normalization; in~\ref{sec:initialization} we give details on how we initialize
 the DNN parameters; in~\ref{sec:sequence_training} we give an overview of how
 we implemented sequence training for DNNs; and in~\ref{sec:online_decoding} we
 discuss online (real-time) decoding using i-vectors for speaker adaptation.

 \subsection{CPU versus GPU-based SGD}
 \label{appendix:cpu:gpu}

  Each machine in our parallel computation implements SGD.
  We have two versions of this, one for GPU and one for CPU.
  The GPU-based computation is standard minibatch-based SGD, typically with 
  512 examples per minibatch.

  In the CPU-based computation, each job uses typically 16 threads in order to
  take advantage of multi-core processors.  The threads share parameters without
  any locks; this is known as Hogwild!~\citep{niu2011hogwild} and was referred
  to in~\citep{Bengio-nnlm2003} as {\em asynchronous}.  In order to prevent
  divergence, each thread processes relatively small minibatches - typically, of
  size 128.

  We should mention at this point that in our formulation, we sum the gradients
  over the elements of the minibatch, rather than averaging: this ensures that
  we make the same amount of progress per sample, regardless of minibatch size,
  and so gives more consistent results when changing the minibatch size.  The need
  to limit the minibatch size in the multithreaded case can be understood as
  follows: think of the effective minibatch size as being the
  minibatch size times the number of threads.  The product of the learning rate
  $\eta$ with the effective minibatch size is relevant for stability of
  the SGD update: if it becomes too large, there is increased danger of divergence.

  We normally use the GPU-based method, because in our experience a GPU can
  process data many times faster than a CPU with multiple threads.  Aside from
  speed, the two methods give very similar results.

 \subsection{Data randomization and sequential data access}
  \label{appendix:data:organize}

 On spinning hard disks, sequential data access can be orders of magnitude more
 efficient than random data access or access to small files.  In the Kaldi
 toolkit~\citep{kaldi_paper}, we try very hard to ensure that any high-volume data access 
 takes the form of sequential reads or writes on large files.

 For neural network training, we keep data access sequential by dumping
 pre-randomized ``training examples'' to disk.  Each training example
 corresponds to a class label together with the corresponding input features,
 including left and right temporal context as needed by the network.  The
 randomization is done just once for the entire data, and the data is accessed in
 the same order on each epoch.  This is probably not ideal from the point of
 view of the convergence of SGD, but our expectation is that for large
 amounts of data the same-order access will not affect the results noticeably.
 
 We break up the training data into $N$ by $M$ rougly equal-sized blocks, where
 $N$ is the number of parallel jobs, specified by the user (typically $4 \leq N
 \leq 8$), and $M \geq 1$ is the number of ``outer iterations per epoch'', which
 is chosen to ensure that the number of samples processed per iteration is close
 to a user-specified value $K$ (e.g. $K = 400\,000$).   The process of randomly
 distributing the data into $N$ by $M$ blocks, and ensuring that the order is randomized
 within each block, is done in parallel; we won't give further details here, because
 the problem is straightforward and there is nothing particularly special about
 our method.  To reduce disk or network access we compress the features on disk
 to 1 byte per float, using a lossy compression method.

 \subsection{Enforcing the maximum parameter change per minibatch}
\label{appendix:weight_limit}

 As mentioned in Section~\ref{sec:max_param_change}, in order to prevent
 instability and parameter divergence we enforce a maximum parameter-change per
 minibatch, which is applied for each layer of the network separately.  Here
 we explain how this is done.  We don't claim that this is an exceptionally good
 method for preventing excessive parameter changes, but we describe it here
 anyway for the sake of completeness.

 Suppose  the update for a single weight matrix is formulated as follows
 (and to keep things simple, we don't include an index for the layer of the network):
\begin{equation}
   \W_{t{+}1} = \W_t + \Delta_t ,
\end{equation}
 where $\Delta_t$ is the change that standard SGD would give us, equal to the
 derivative of the objective function for this minibatch multiplied by the learning rate $\eta_t$. 
 To enforce the maximum parameter chanbge, we scale the change by a scalar $\alpha_t$:
\begin{equation}
 \label{eqn:add:w}
   \W_{t{+}1} = \W_t + \alpha_t \Delta_t ,
\end{equation}
 where we would like to choose $\alpha_t \leq 1$ to ensure that $||\alpha_t
 \Delta_t||_F$ does not exceed a specified limit, $||\cdot||_F$ being the
 Frobenius norm.  However, we don't implement this scheme exactly as described
 above because it would involve creating a temporary matrix to store the product
 of matrices $\Delta_t$ just in order to compute its norm, and we don't want
 to incur this penalty.

 Instead we enforce it in a way that involves a sum over elements of the minibatch.
 If $\Delta_t = \eta \X^T \Y$, then $\Delta_t$ can be written as a sum over an index $i$ that
 ranges over the rows of $\X$ and $\Y$.  By properties of norms,
 the 2-norm of $\Delta_t$ cannot exceed the sum of the 2-norms of the terms in
 this sum: if the rows of $\X$ and $\Y$ are written as $\x_i$ and $\y_i$, then
\begin{equation}
  ||\Delta_t||_F \leq \sum_i \eta ||\x_i||_2 ||\y_i||_2  \label{eqn:delta_t_approx}
\end{equation} 
 It does not take excessive time or memory to compute the vector norms $||\x_i||_2$ and $||\y_i||_2$,
 so we compute the right hand side of~\ref{eqn:delta_t_approx} and use it as a stand-in
 for $||\Delta_t||_F$, giving us
\begin{equation}
  \alpha_t = \min\left( 1,  \frac{ \mbox{max-change-per-minibatch} } { \sum_i \eta ||\x_i||_2 ||\y_i||_2 } \right)
\end{equation}
 where $\mbox{max-change-per-minibatch}$ is a user-specified maximum parameter-change per minibatch.
 Empirically we have found that it tends to be necessary to increase $\mbox{max-change-per-minibatch}$ when
 using a larger minibatch size, so to simplify the configuration process we define
\begin{equation}
 \mbox{max-change-per-minibatch} = N \mbox{max-change-per-sample}
\end{equation}
 where $N$ is the minibatch size.  We always set $\mbox{max-change-per-sample}$ to 0.075 for
 experiments reported here.  To clarify how this method interacts with the natural gradient
 methods described in Section~\ref{sec:natural:gradient}: the natural gradient is implemented as a
 modification to the $\X$ and $\Y$ matrices, so we simply apply this maximum-change logic on top of the modified
 $\X$ and $\Y$ quantitities. 

 What we've found that this maximum-parameter-change limit is active only early in training
 for layers closer to the output.

 \subsection{Generalized model averaging}
 \label{sec:model_averaging}

 The convergence theory of Stochastic Gradient
 Descent~\citep{kushner2003stochastic} suggests that, for convex
 problems, if we take not the last iteration's model parameters but the average
 over all iterations, it can improve the convergence rate, particularly in
 `poorly-conditioned' problems (i.e. where the condition number of the Hessian
 is very large).  This is not applicable in non-convex problems such as ours,
 but it does suggest a
 related method.  As mentioned above, we define an {\em outer iteration} as the length of time
 it takes for all jobs to process $K$ samples (e.g. $K = 400\,000$), and on each outer iteration each job
 dumps its final model to disk and we average these to produce a single model.  We store the models
 (averaged over all jobs) for each outer iteration.  At the very end of training,
 instead of choosing the model from the
 final outer iteration, we take the models from the last $P$ outer iterations (e.g. $P = 20$),
 and search for a generalized weighted combination of these models that optimizes
 the objective function on a subset of training data-- we tried using validation data here, but for
 our task we found it worked best to use training data.  By {\em generalized
 weighted combination}, what we mean is that the parameters are a weighted combination of
 the parameters of the input models, but each layer can have different weighting factors.  Thus, if there are $P$
 models and $L$ layers, the number of parameters we learn on the data subdset is $LP$.  A few more details:
\begin{itemize}
    \item The optimization method is L-BFGS.
    \item To improve the convergence speed of L-BFGS, we optimize in a transformed (preconditioned) space where
       the preconditioner is related to the Fisher matrix.
    \item The starting point for the optimization is the best of $P + 1$ choices, corresponding to each of
       the $P$ final iterations, and the average of all of them.
    \item We add a very tiny regularizer (like $10^{-10}$ times the square of the vector of weights)
       to stop the weights going to infinity in cases (like p-norm networks) where the objective function is
       invariant to the parameter scale.
    \item We generally aim to optimize over the last $P = 20$ models (assuming they share the same parameter
       structure, e.g. we haven't added layers).
    \item In cases where $P$ iterations would amount to less than one epoch, we optimize over $P$ models where
       the individual models are simple averages of model parameters for a duration of about $1/P$ of the entire
       epoch.
\end{itemize}
 We have generally found that this model combination slightly improves results, but it 
 is not a focus of the current paper so we don't provide experimental results for
 this here.

 \subsection{Mixture components (sub-classes)}
 \label{sec:mixture_components}

 When using Gaussians for speech recognition, the usual approach is to use a Gaussian mixture model (GMM)
 rather than a single Gaussian, to model each speech state.  We have generalized this idea to neural
 networks, by allowing the posterior of each speech state to be written as a sum over the posterior
 of ``sub-classes'' that are analogous to the Gaussians in a GMM.
 About halfway through training, we ``mix up'' the model by increasing the 
 dimension of the softmax layer to a user-specified
 number that is greater than the number of classes (usually about double the number of classes).
 After the softmax layer we introduce a ``sum-group'' layer which sums its input over fixed groups of indexes
 to produce a posterior for each class that is a sum over the posteriors of the hidden ``sub-classes''.
 We also tried sharing the sub-classes across classes in groups, but did not find this helpful.

 Rather than distributing the ``sub-classes'' evenly, we allocate more
 sub-classes to the more common classes.  We allocate them proportional to the
 $1/3$ power of the count of that class in the training data; this is based on
 the rule we use to allocate Gaussians in our GMMs.

 When initializing the parameters of the ``mixed-up'' final weight matrix, we
 make it correspond quite closely with the original weight matrix.  Each row of
 the new weight matrix corresponds to a row of the old weight matrix, plus a
 small noise term to allow the values of the rows to diverge; and we modify the
 bias term to normalize for the fact that some classes have more sub-classes
 than others.

 We have generally found that this slightly improves results, but again, this is
 not a focus of the current paper and we won't be showing experimental results
 about this.  More recent experimental results\footnote{Thanks to Minhua Wu}
 show that the mixture components may not have diverged sufficiently that we can
 regard them as truly distinct; the method may only be helping because it has
 the same effect as decreasing the learning rate for the final-layer parameters
 corresponding to the higher-frequency classes.  We describe it only for
 completeness.

 {\em This note is being added after publication as an ICLR workshop paper.
   Further experiments conducted by Minhua Wu discovered that the mixture
   components were not diverging sufficiently to be regarded as distinct
   mixtures, and the observed small improvement was likely due to a stabilizing
   effect on the update of parameters for higher-count classes, by splitting the
   gradient into multiple pieces.  We were able to improve our results slightly
   by removing the mixture-component and adding instead a fixed scaling as a
   separate component/layer after the final weight matrix and before the
   softmax, with scales proportional to (data-count)$^{-0.25}$, renormalized so
   that the average was 1.  }

 \subsection{Input data normalization}
 \label{sec:input_data_normalize}

 As mentioned in~\citep[Section 4.3]{lecun2012efficient}, when training neural
 networks it is helpful to normalize the input data so that it is zero mean and
 so that more important dimensions of input data have a larger variance.  We
 wanted a generic way to achieve this that would be invariant to arbitrary
 affine transforms of the input.  The technique we developed requires as
 statistics a within-class covariance $\W$ and a between-class covariance $\B$,
 accumulated from the class-labeled data as if in preparation for multi-class
 Linear Discriminant Analysis (LDA).  Assume in what follows that we have
 already normalized the data so that it is zero-mean.  For the technique we are
 about to describe to make sense, the number of classes should not be much
 smaller than the feature dimension; fortunately, in our case it is much
 larger-- $5000 > 300$, to give typical numbers.

 Suppose we were to do multi-class LDA but not actually reduce the dimension.
 We would transform into a space where $\W$ was unit and $\B$ was diagonalized.  Suppose
 the $\B$ in this space has diagonal elements $b_i$.  Then the total covariance
 in each dimension $i$ is $b_i + 1$.  This has the desirable property that the data covariance
 is higher in ``more important'' directions, but it doesn't drop as fast
 as we'd like for unimportant directions-- it never goes below 1.  In our method,
 we do the LDA-type transform as mentioned above, then scale each row of the transform
 by $\sqrt{ (b_i + 0.001) / (b_i + 1) }$.  After this scaling, the total covariance becomes
 $b_i + 0.001$, where $b_i$ is the ratio of between-class to within-class covariance.
 This seems to work well.
 
 After creating the transform matrix as described above, we do a singular value
 decomposition on it, floor the singular values (normally to 5), and reconstruct
 again.  The motivation here is to avoid a rarely encountered pathology that
 occurs when the training data covariance was close to singular, which leads to
 a transform with very large elements, that might produce very large transformed
 data values on mismatched test data or due to roundoff.  This step rarely
 floors more than a handful of singular values so has little effect on the
 transform.

 \subsection{Parameter initialization}
 \label{sec:initialization}

 We decided not to implement generative pre-training as
 in~\citep{hinton2006fast}, because while it is well established that it improves
 results for small datasets, our understanding is that as the amount of training
 data gets larger, it eventually gives no improvement compared to a suitable
 random initialization or discriminative layer-wise backpropagation as
 in~\citep{seide2011feature}.  We could not find a published reference for this;
 it is something we have been told verbally.  We refer here specifically to
 speech recognition tasks; this does not apply to tasks like computer vision
 where much larger networks are used.  In fact, the alternative ``nnet1''
 implmentation of DNNs in Kaldi does support pre-training, and for small
 datasets (say, 50 hours or less), it generally gives slightly better results
 than the ``nnet2'' implementation which we speak of here.  For larger datasets,
 the ``nnet1'' implementation eventually becomes impractical to run because it
 takes too long, and a detailed comparison is way beyond the scope of this
 paper.

 Instead of pre-training, we use what is described in~\citep{seide2011feature} as
 layer-wise back-propagation (BP).  What this means is, we initialize a network
 with one hidden layer, train with BP for a short time (two ``outer iterations''
 for our experiments reported here), then remove the final softmax layer and add
 a new, randomly initialized hidden layer on top of the existing hidden layer;
 train for a short time again; and repeat the process until we have the desired
 number of hidden layers.  Similar to~\citep{glorot2010understanding}, we use a
 standard deviation of $\frac{1}{\sqrt{i}}$ for the weights, where $i$ is the
 fan-in to the weight matrix; but we initialize the parameters of softmax layers
 to zero.  Note: we found it essential to discard the parameters of the final
 softmax layer when adding each new hidden layer, as prescribed
 in~\citep{seide2011feature}.

 For smaller datasets we can improve results versus layer-wise BP by
 initializing all but the last layer of the network from a network trained on
 another large dataset, possibly from another language.  When initializing this
 way we typically find it best to use a larger network than we otherwise would
 have used.

 Because we noticed that sometimes on an outer iteration immediately following
 the random initialization of parameters (including the first outer iteration),
 the parameter averaging can degrade rather than improve the objective function,
 we modified our parallel training method so that on these iterations, instead
 of averaging the parameters we choose the one that had the best objective
 function computed on the subset of data that it was trained on (this
 approach avoids any extra computation).

 \subsection{Sequence training}
 \label{sec:sequence_training}

 Sequence training~\citep{mohamed2010investigation} is a term that has the the
 same meaning for DNNs that ``discriminative training'' \citep{thesis} has in the
 speech recognition community for GMMs.  It is a collective term for various
 objective functions used for training DNNs for sequence tasks, that only make
 sense at the whole-sequence level.  This contrasts with the cross-entropy
 objective function which, given a fixed Viterbi alignment of the HMM states,
 easily decomposes over the frames of training data.  In GMM-based speech
 recognition, the term ``discriminative training'' contrasts with Maximum
 Likelihood estimation; in DNN-based speech recognition it contrasts with
 cross-entropy training.  There are two popular classes of
 sequence/discriminative objective functions:
\begin{itemize}
    \item Maximum Mutual Information (MMI)-like objective functions~\citep{bahl1986maximum,thesis}, 
        more properly called conditional maximum likelihood: these have the form of sum over all utterances of
        the log-posterior of the correct word sequence for each utterance, given the model and the data.
        These include its popular 'boosted' variant~\citep{poveyboostmmi} which is inspired by margin-based
        objective functions.
    \item Minimum Bayes Risk (MBR)-like objective functions: popular variants
        include Minimum Phone Error (MPE)~\citep{mpe,thesis} and state-level Minimum Bayes Risk~\citep{smbr,smbr_dan}.  These
        have the form of an expectation, given the data and the model, of an
        edit-distance type of error.  We can compute its derivative w.r.t. the model parameters, because
        the posteriors of the different sequences vary with the model parameters.
\end{itemize}
This paper is mainly about our parallel approach to standard cross-entropy training.  However, we
also apply the same ideas (model averaging, NG-SGD) to sequence training.
We generally use state-level Minimum Bayes Risk 
(sMBR)~\citep{smbr,smbr_dan} although we have also implemented Minimum Phone Error (MPE)~\citep{mpe} 
and Boosted MMI~\citep{poveyboostmmi}.  The high-level details of our lattice-based
training procedure are similar to~\citep{sequence_dnn}, but  note that in that paper we
describe an alternative implementation of deep neural nets (the ``nnet1'' setup) that exists
within Kaldi; this paper is about the alternative ``nnet2'' setup.  Some items in common with 
the sequence training described in that paper include the following:
\begin{itemize}
  \item We use a low, fixed learning rate (e.g. 0.00002).
  \item We generally train for about 4 epochs.
\end{itemize}
Some differences include the following:
\begin{itemize}
  \item We do parallel SGD on multiple machines, with periodic model averaging.
  \item Rather than randomizing at the utterance level, we split the lattice 
       into as small pieces as possible given the lattice topology, and excise parts of the lattice that would not
       contribute nonzero derivatives; and we randomize the order of the remaining pieces.
  \item To ensure that all layers of the network are trained about the same amount,
      we modify the learning rates in order to ensure that the relative change
      in parameters on each ``outer iteration'' is the same for each layer; their
      geometric average is constrained to equal the user-specified fixed learning rate (e.g. 0.00002) which we
      mentioned above.
  \item In our recipe, we generate the lattices only once.
  \item The minibatches actually consist of several small chunks of lattice (split as described
    above), from many different utterances, spliced together.
\end{itemize}
Something that we should note in connection with the learning rates
is that for p-norm networks, since the network output is
invariant to (nonzero) scaling of the parameters of the p-norm layers\footnote{
This is thanks to the ``renormalization layers'' that follow each p-norm layer~\citep{zhang2014improving}}, and 
since the generalized weighted combination of Section~\ref{sec:model_averaging} may output arbitrarily 
scaled weights, it is hard to specify in advance a suitable learning rate.  To
solve this problem, we first scale the parameters of p-norm layers so
so that the expected square of a randomly chosen matrix element is one.

For sequence training, because the frames in a minibatch are not drawn
independently from the training data but consist of sequential frames from one
or a few utterances, our ``simple'' NG-SGD method is not applicable,
and we only apply the online method.



 \subsection{Online decoding and i-vector inputs}
 \label{sec:online_decoding}

 In speech recognition applications it is sometimes necessary to process data continuously
 as it arrives, so that there will be no latency in response.  This makes it necessary
 that the algorithms used should not have any dependencies that are ``backwards'' in time.
 Backwards-in-time dependencies in our conventional neural net recipes, e.g. as
 reported in~\citep{zhang2014improving}, include cepstral mean normalization (CMN),
 in which we subtract the mean of the input features; and fMLLR adaptation, also
 known as constrained MLLR adaptation~\citep{orig_cmllr}, in which
 we use a baseline GMM system to compute a likelihood-maximizing linear transform of the
 features.  Although we use ``online'' versions of both of these things for online GMM-based
 decoding, it makes the system very complex and is not ideal for combination with DNNs.

 In order to have a system that is easier to turn into an online algorithm, we
 use i-vectors~\citep{dehak2011front} as an additional input to the neural
 network, in addition to the spliced cepstral features.  This has been done
 before, e.g.~\cite{saon2013speaker,bacchiani2013rapid}.  An i-vector is a vector
 normally of dimension in the range of several hundred, that represents speaker
 characteristics in a form suitable for speaker identification, and which is
 extracted in a Maximum Likelihood way in conjunction with a single
 mixture-of-Gaussians model (their means are regressed on the i-vector).  The
 parameters of the factor analysis model that extracts the i-vectors are trained
 without any supervision, just on a large number of audio recordings.  In our
 case we extract i-vectors of dimension 100.  Once the i-vector extractor is
 trained, we switch to ``online'' extraction of i-vectors for both training and
 decoding, in which only frames preceding the current frame are taken as inputs
 to the i-vector estimation process.  At the beginning of the utterance, the
 i-vector will be zero due to the prior term.

 The actual inputs to the DNN in this setup normally consist of the i-vector,
 plus $\pm 7$ frames of Mel frequency cepstral coefficients
 (MFCCs)~\citep{davis1980comparison}, without cepstral mean normalization.  Some
 other authors~\citep{senior2014improving} use log Mel filterbank energies; the
 MFCC features we use here are equivalent to log Mel filterbank energies because
 MFCCs are a linear transform of them (we use the same number of coefficients as
 filterbanks, 40 for these experiments) and our input data normalization
 (Section~\ref{sec:input_data_normalize}) is invariant to such transforms; we
 only use MFCCs because they are more easily compressible and our ``training
 example'' data structure (Appendix~\ref{appendix:data:organize}) compresses the input
 features.

 In order to train models that are well matched both to per-speaker decoding,
 where statistics from previous utterances of the same speaker are included in
 the i-vector estimation, and per-utterance decoding, where we make a fresh
 start each time, we generally train after splitting the speakers into ``fake''
 speakers that each have no more than two utterances.

 In experiments on a number of datasets, we have generally found that this method
 gives us about the same performance as our previous recipe where we trained
 a DNN on top of ${\pm 4}$ frames of the standard 40-dimensional features
 consisting of mean-normalized MFCC features processed with LDA and MLLT, and speaker
 adapted with fMLLR (a.k.a. constrained MLLR~\citep{orig_cmllr}).  We prefer it
 due to its convenience for applications and its convenience for 
 cross-system transfer learning.

\end{document}